\documentclass[acmtog]{acmart}

\AtBeginDocument{%
  \providecommand\BibTeX{{%
    \normalfont B\kern-0.5em{\scshape i\kern-0.25em b}\kern-0.8em\TeX}}}
\usepackage{enumitem}
\usepackage{color}
\definecolor{applegreen}{rgb}{0.55, 0.71, 0.0}
\definecolor{autumnorange}{rgb}{0.87, 0.61, 0.33}
\definecolor{creampurple}{rgb}{0.68,0.53,0.78}

\newcommand{\revise}[1]{{\color{black}{#1}}}

\usepackage{soul}
\setcopyright{acmlicensed}
\copyrightyear{2024}

\setcopyright{acmlicensed}
\acmJournal{TOG}
\acmYear{2024} \acmVolume{43} \acmNumber{6} \acmArticle{212} \acmMonth{12}\acmDOI{10.1145/3687931}

\acmISBN{978-1-4503-XXXX-X/18/06}
\acmSubmissionID{507}
\begin{document}

\title{StyleTex: Style Image-Guided Texture Generation for 3D Models}
\author{Zhiyu Xie}
\authornote{Equal contribution}
\affiliation{
  \institution{State Key Lab of CAD\&CG, Zhejiang University}
  \city{Hangzhou}
  \state{Zhejiang}
  \country{China}
}
\email{xiezhiyu@zju.edu.cn}

\author{Yuqing Zhang}
\authornotemark[1]
\affiliation{
  \institution{State Key Lab of CAD\&CG, Zhejiang University}
  \city{Hangzhou}
  \state{Zhejiang}
  \country{China}
}
\email{3180102110@zju.edu.cn}

\author{Xiangjun Tang}
\affiliation{
  \institution{State Key Lab of CAD\&CG, Zhejiang University}
  \city{Hangzhou}
  \state{Zhejiang}
  \country{China}
}
\email{xiangjun.tang@outlook.com}

\author{Yiqian Wu}
\affiliation{
  \institution{State Key Lab of CAD\&CG, Zhejiang University}
  \city{Hangzhou}
  \state{Zhejiang}
  \country{China}
}
\email{onethousand1250@gmail.com}

\author{Dehan Chen}
\affiliation{
  \institution{State Key Lab of CAD\&CG, Zhejiang University}
  \city{Hangzhou}
  \state{Zhejiang}
  \country{China}
}
\email{cdh573885@outlook.com}

\author{Gongsheng Li}
\affiliation{
  \institution{Zhejiang University}
  \city{Hangzhou}
  \state{Zhejiang}
  \country{China}
}
\email{ligongshengzju@foxmail.com}

\author{Xiaogang Jin}
\authornote{Corresponding author.}
\affiliation{
  \institution{State Key Lab of CAD\&CG, Zhejiang University}
  \city{Hangzhou}
  \state{Zhejiang}
  \country{China}
}
\email{jin@cad.zju.edu.cn}


\begin{abstract}
Style-guided texture generation aims to generate a texture that is harmonious with both the style of the reference image and the geometry of the input mesh, given a reference style image and a 3D mesh with its text description.  
Although diffusion-based 3D texture generation methods, such as distillation sampling, have numerous promising applications in stylized games and films, it requires addressing two challenges: 1) decouple style and content completely from the reference image for 3D models, and 2) align the generated texture with the color tone, style of the reference image, and the given text prompt.
To this end, we introduce StyleTex, an innovative diffusion-model-based framework for creating stylized textures for 3D models. Our key insight is to decouple style information from the reference image while disregarding content in diffusion-based distillation sampling.
Specifically, given a reference image, we first decompose its style feature from the image CLIP embedding by subtracting the embedding's orthogonal projection in the direction of the content feature, which is represented by a text CLIP embedding. 
Our novel approach to disentangling the reference image's style and content information allows us to generate distinct style and content features. 
We then inject the style feature into the cross-attention mechanism to incorporate it into the generation process, while utilizing the content feature as a negative prompt to further dissociate content information. 
Finally, we incorporate these strategies into StyleTex to obtain stylized textures. We utilize Interval Score Matching to address over-smoothness and over-saturation, in combination with a geometry-aware ControlNet that ensures consistent geometry throughout the generative process. The resulting textures generated by StyleTex retain the style of the reference image, while also aligning with the text prompts and intrinsic details of the given 3D mesh.
Quantitative and qualitative experiments show that our method outperforms existing baseline methods by a significant margin.
\end{abstract}


\begin{teaserfigure}
  \includegraphics[width=\textwidth]{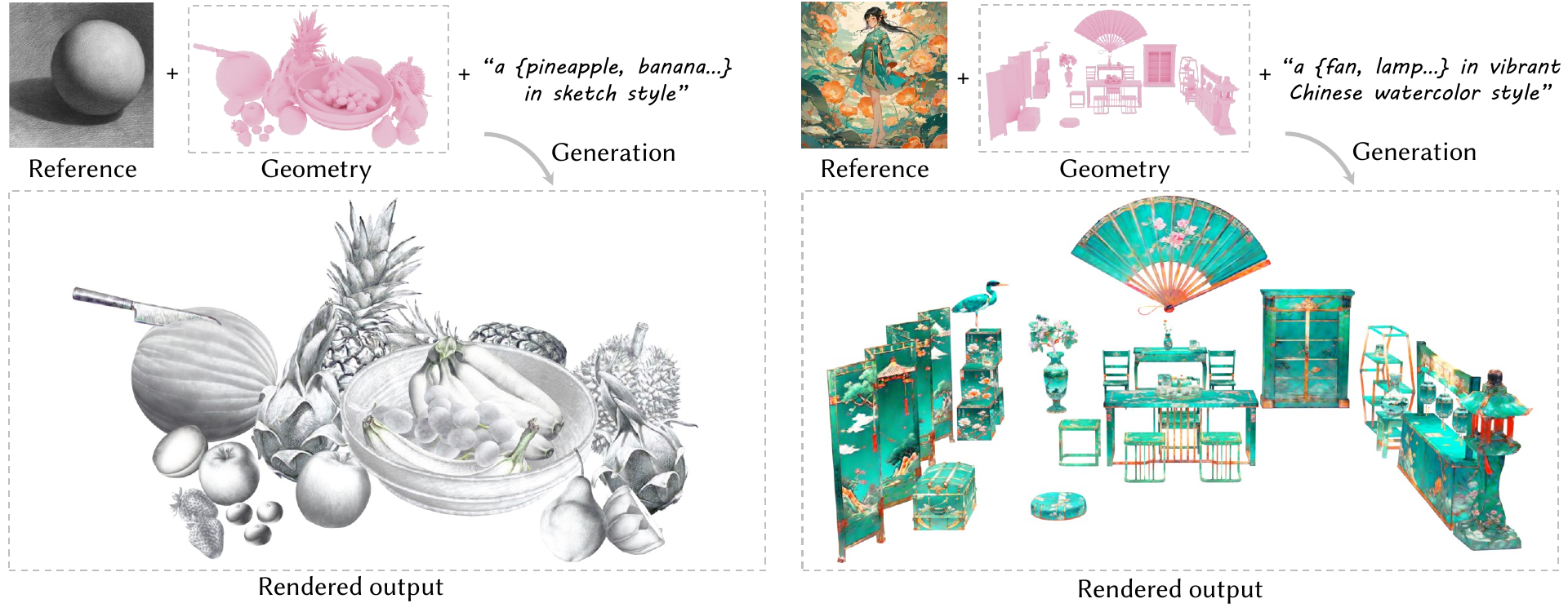}
  \vspace{-5mm}
  \caption{
  StyleTex is capable of generating visually compelling and harmonious stylized textures for a given scene. 
  For each mesh in the 3D scene, StyleTex utilizes the untextured mesh, a single reference image, and a text prompt describing
  the mesh and desired style as inputs to generate a stylized texture.
  The generated textures preserve the style of the reference image while ensuring consistency with both the text prompts and the intrinsic details of the given 3D mesh. 
  At the bottom, we present the rendered output for the provided 3D scene with the generated texture.
  }
  \label{fig:teaser}
\end{teaserfigure}
\vspace{-3mm}
\maketitle
\section{Introduction}
\label{sec:intro}

We investigate an under-explored generation problem: style image-guided texture synthesis, which is crucial in computer vision and graphics, facilitating the creation of visually compelling and immersive digital environments in games and films. The generated texture needs to be harmonious with both the 3D shape and style of the reference image, which requires the texture to align with the geometry while conveying a consistent style from different views.

Existing research mostly investigates the above two requirements separately. In 2D style-image generation methods, the style is conveyed by separating it from the reference image and incorporating it into the final output, which usually involves fine-tuning~\cite{hu2022lora,gal2022textual,ruiz2022dreambooth} the diffusion model to be a stylized image generator or adjusting the hidden layers of the diffusion model with the extracted style features~\cite{jeong2024visual,hertz2024style,wang2024instantstyle, voynov2023p+, he2024freestyle}. 
In parallel, 3D texture can be generated by iteratively inpainting~\cite{richardson2023texture, chen2023text2tex} or image synthesis with multi-view consistency ~\cite{cao2023texfusion,liu2023text,gao2024genesistex,wu2024texro}. 
More recently, distillation methods such as score distillation sampling~\cite{metzer2023latent,Chen_2023_ICCV,youwang2023paint} have also proven their superior effectiveness in synthesizing 3D consistent textures. 
%
Compared to the direct generation of textures, distillation methods are capable of achieving better view and global style consistency while avoiding local seam problems. 

Despite the progress in these two distinct areas, incorporating the desired style into texture generation is not straightforward. 
One possible solution is to combine the distillation method with a diffusion distribution aligned with the reference image's style. 
However, this leads to two challenges: 1) decoupling the style and content from the reference image entirely, and 2) preserving the color tone. Firstly, the ambiguity between style and content from different views complicates the decoupling process. In 2D domains, separating style and content within a single viewpoint may succeed in most situations. 
However, in 3D domains, failure to effectively decouple style from any single viewpoint can result in inaccurate style and unintended content leakage in the final texture. Thus, the generation of stylized textures in 3D domains requires a robust method for disentangling style and content.
Secondly, distillation methods may result in over-saturation and over-smoothing within the generated textures, leading to color shifts and a lack of details,  
hindering the accurate reflection of the intended style.

To overcome these challenges, we propose \textbf{StyleTex}, a diffusion-model-based pipeline to generate style textures under the guidance of a single image. 
Our key insight is to extract the style information from the reference image while disregarding the content information.
Inspired by the multi-modal applications of the CLIP space, we propose to represent the content of the reference image as the CLIP embedding of its corresponding text prompt.
A naive method to discard the content from the reference image \revise{in InstantStyle~\cite{wang2024instantstyle}} is to drive the reference image embedding in the same CLIP space toward the opposite direction of the content embedding. However, the slight misalignment between the content embedding and the real content information of the image may cause undesirable image embedding alerting, which results in unclean content information remaining or color tone changing. 
To address this, we remove the content information from the reference image embedding by decomposing its CLIP embedding into two separate orthogonal features. 
One of these features aligns with the content embedding and encodes most of the content information of the reference image. We retain only the remaining feature, which predominantly relates to the style, to refine our diffusion model.
To this end, we explicitly incorporate the style-relevant feature through the cross-attention mechanism, which also serves as a color tone guidance that can prevent unintentional color tone changing during the distillation process. Furthermore, we incorporate the content embedding as a negative prompt to further dissociate content information. 
We integrate the aforementioned strategies into StyleTex to generate stylized textures and utilize Interval Score Matching (ISM) \cite{EnVision2023luciddreamer} to further tackle the issue of over-smoothness. Moreover, we utilize a geometry-aware ControlNet to ensure geometric consistency throughout the generative process.

In summary, our work makes the following major contributions:
\begin{itemize}
    \item  A diffusion-model-based pipeline to generate style textures under the guidance of a single image, enabling the automatic creation of diverse stylized virtual environments.

    \item A novel style decoupling and injection strategy that effectively guides stylization while addressing issues of content leakage and style deviation in texture generation. 
    
\end{itemize}
\section{Related Work}

\subsection{Image guided stylization}

Given a reference image, image-guided stylization aims to synthesize a new image that shares the same style as the reference image while demonstrating the intended content. Early methods~\cite{gatys2016imagea, Chen2016FastPS, gu2018arbitrary} alter the style of an image while preserving its content by solving a slow optimization. The following methods propose to represent the style by a neural network~\cite{artflow2021, Ulyanov2016TextureNF, dumoulin2017alearned, chen2017stylebank, johnson2016perceptual, zhng2019multistyle}, or by the statistics of the hidden features of a network~\cite{huang2017arbitrary, WCT-NIPS-2017, park2019arbitrary, kotovenko2019iccv, kolkin2022neural}, enabling stylization by a single-step inference of the network. With the development of the text-to-image diffusion model~\cite{rombach2021highresolution}, \revise{fine-tuning the diffusion model~\cite{hu2022lora,frenkel2024blora, sohn2023styledrop,shah2023ZipLoRA,  ruiz2022dreambooth,chen2023controlstyle, gal2022textual} yields a stylized image generator but requires time-consuming training. 
%
Based on the existing style representations, 
modifying the structure of the diffusion model  ~\cite{Zhang2023inst,hertz2024style, jeong2024visual} and utilizing other adapter-based methods~\cite{ye2023ip-adapter, wang2024instantstyle,ye2023styleadapter,qi2024deadiff} } allows for the incorporation of desired styles without training.

Image-guided 3D stylization can be analogous to the 2D methods but replaces the 2D image with the 3D representations, such as point clouds~\cite{huang_2021_3D_scene_stylization, mu20213D}, \revise{mesh~\cite{kato2018renderer, yin2021_3DStyleNet,hollein2022stylemesh},  NeRF~\cite{kolkin2022neural, huang2022stylizednerf, zhang2022arf, nguyen-phuoc2022snerf,liu2023stylerf} or 3D Gaussian~\cite{Zhang2024StylizedGS}. }
However, establishing style consistency over multiple views in 3D space has not been fully explored, leading to artifacts such as content leakage.

\subsection{Text/Image-guided Texture Generation}
Automatically generating textures over 3D surfaces has garnered widespread attention and important applications. 
While training the texture generation network on a small dataset ~\cite{chen2022AUVNET,siddiqui2022texturify} aids in learning a stylized distribution, it also restricts the network to a particular texture category.
Text-to-image diffusion ~\cite{rombach2021highresolution} incorporates a strong 2D image prior that represents a real image distribution, offering robust guidance for text-driven texture generation.
For instance,
TEXTure~\cite{richardson2023texture} and Text2Tex~\cite{chen2023text2tex} employ the diffusion model to iteratively inpaint the geometry from different viewpoints. However, the 2D diffusion model lacks an understanding of 3D shape and multi-view color consistency, leading to blurry and low-quality texture results.
%
To maintain 3D consistency, a possible way is to employ a 3D consistent prior~\cite{le2023euclidreamer,Chen_2023_ICCV,metzer2023latent,guo2024decorate3D}, such as applying the score distillation sampling using a geometry-conditioned diffusion model. In addition, methods such as SyncMVD~\cite{liu2023text}, TexRO~\cite{wu2024texro} and GensisTex~\cite{gao2024genesistex} are also able to maintain the 3D consistency by explicitly projecting the intermediate results of each denoising step into a consistent texture space.

%
Instead of employing diffusion models designed for a real-image distribution, another viable alternative could be to fine-tune a diffusion model to learn a UV space texture distribution~\cite{zeng2023paint3D,liu2024texdreamer,Geometry_Aware_Texturing}. This approach can significantly accelerate the generation process, but the results may be heavily impacted by the quality of the UV mapping.
%

Unlike text-driven texture generation, image-guided approaches require interpreting the style of an image and hence cannot simply rely on the pretrained text-to-image model. 
\revise{In addition to text-guided texture generation, there have also been attempts in image-guided texture generation. TEXTure~\cite{richardson2023texture} employs textual inversion~\cite{gal2022textual} to capture the style and structural features of reference images, while Texturedreamer~\cite{yeh2024texturedreamer} uses Dreambooth~\cite{ruiz2022dreambooth} to fine-tune the Stable Diffusion and then applies the personalized model in the geometry-aware score distillation.}
%
However, these methods often require a fine-tuning process and cannot exclude the content information of the reference images. 
In contrast, our method is dedicated to decoupling the style and content information and generating textures consistent with the style of the reference images. As a result, the content and details of the textures are consistent with the textual prompts and the model's geometry, all without the need for an additional training process.
\section{Method}
\begin{figure*}
  \includegraphics[width=\textwidth]{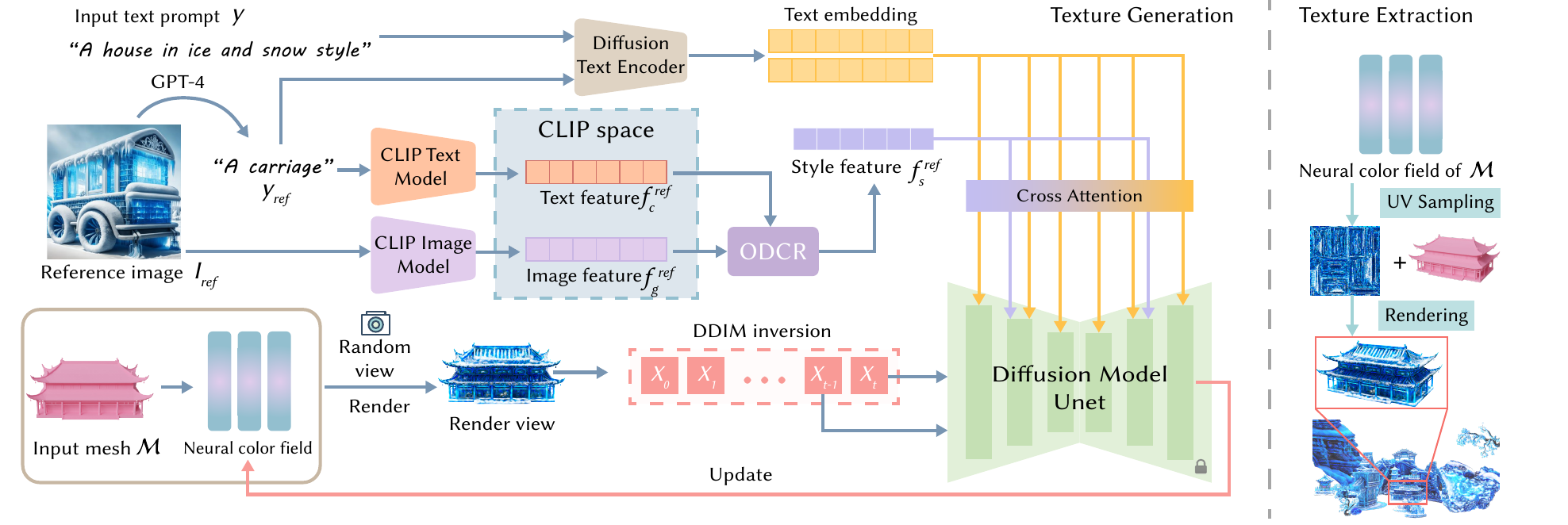}
  \vspace{-4mm}
  \caption{\textbf{Overview of our pipeline. }
  StyleTex's inputs include a reference style image $I_{ref}$, a text prompt $y$, and an untextured 3D mesh $\mathcal{M}$.
  During training, we utilize our innovative ODCR method (described in Sec. \ref{sec: ism}) to extract a content-unrelated style feature, $f_s^{ref}$, from the reference image. The style feature and text embeddings are fed into the Unet to guide the optimization of the texture field.
  During inference, texture maps can be sampled from the texture field and directly employed in downstream game or film production, enabling the creation of stylized digital environments.
  }
  \label{fig:pipeline}
  \vspace{-4mm}
\end{figure*}
Given an untextured mesh, a textual prompt, and a reference image, our goal is to generate textures consistent with the image style while aligning the content of the textures with both the textual prompt and the geometry of the model. In Sec.~\ref{sec:preliminary}, we introduce the prior knowledge relevant to our method, including the diffusion denoising process and interval score matching (ISM) loss. 
In Sec.~\ref{sec: generation}, we present our pipeline for generating stylized textures. 
In Sec.~\ref{sec: ism}, we present our approach to style infusion, which includes transformer layer style injection and content and style disentanglement.

\subsection{Preliminary}
\label{sec:preliminary}
When it comes to text-to-3D, numerous approaches have been developed to optimize 3D representations by distilling 2D diffusion models, using techniques like score distillation sampling (SDS) \cite{poole2022dreamfusion}. The optimization goal of SDS is to make the renderings of 3D representations align with the image distribution in a pre-trained text-to-image diffusion model.  
At each iteration, the differentiable rendering function $g$ renders the \revise{trainable paramaters }$\theta$ from camera $c$, getting the rendered image $x_0$. After that, $x_0$
undergoes a noise addition process, resulting in $x_t \sim \mathcal{N}\left(x_t ; \sqrt{\bar{\alpha}_t} x_0,\left(1-\bar{\alpha}_t\right) \boldsymbol{I}\right)$. With a text prompt $y$, a pre-trained 2D diffusion model is utilized to predict the corresponding noise. The gradient of the SDS loss with respect to the 3D representation is determined as follows:
\revise{
\begin{equation}
    \nabla_\theta \mathcal{L}_{\mathrm{SDS}}(\theta) \approx \mathbb{E}_{t, \epsilon, c}\left[\omega(t)\left(\epsilon_\phi\left(x_t, t, y\right)-\epsilon\right) \frac{\partial g(\theta, c)}{\partial \theta}\right],
\end{equation}
where $\epsilon \sim \mathcal{N}(\mathbf{0}, \boldsymbol{I})$ is the ground truth denoising direction of $x_t$ at timestep $t$, $\epsilon_\phi\left(x_t, t, y\right)$ is the predicted denoising direction~\cite{EnVision2023luciddreamer} under the given condition $y$, }and $\omega(t)$ denotes a weighting function that absorbs the constant $\alpha_t \textbf{I} = \partial x_t / \partial x_0$. 
This equation can be rewritten as:
\begin{equation}
    \nabla_\theta \mathcal{L}_{\mathrm{SDS}}(\theta)=\mathbb{E}_{t, \epsilon, c}\left[\frac{\omega(t)}{\gamma(t)}\left(x_0-\hat{x}_0^t\right) \frac{\partial g(\theta, c)}{\partial \theta}\right],
\end{equation}
\revise{where $\gamma(t)= \frac{\sqrt{1-\bar{\alpha}_t}}{\sqrt{\bar{\alpha}_t}}$, and $\hat{x}_0^t=\frac{x_t-\sqrt{1-\bar{\alpha}_t} \epsilon_\phi\left(x_t, t, y\right)}{\sqrt{\bar{\alpha}_t}}$ is the pseudo-GT~\cite{EnVision2023luciddreamer} estimated by the single-step Diffusion Probabilistic Model (DDPM)~\cite{ho2020denoising} .}

Based on SDS, Interval Score Matching (ISM) \cite{EnVision2023luciddreamer} generates a reversible diffusion trajectory by adding noise to $x_0$ through Denoising Diffusion Implicit Models (DDIM)~\cite{song2020denoising} inversion, and employing multi-step DDIM denoising process. This helps to achieve a more consistent and higher-quality $\hat{x}_0^t$. This process of noise addition and subsequent denoising facilitates the neutralization of a series of neighboring interval scores with opposing scales, resulting in the formulation of the ISM loss:
\begin{equation}
    \begin{aligned}
       \nabla_\theta \mathcal{L}_{\mathrm{ISM}}(\theta)= \mathbb{E}_{t, c}\left[\omega(t) \delta(x_t,x_{t-1},t,t-1)\frac{\partial g(\theta, c)}{\partial \theta}\right],
    \end{aligned}
\end{equation}
\revise{\begin{equation}
\delta(x_t,x_{t-1},t,t-1)=\epsilon_\phi\left(x_t, t, y\right)-\epsilon_\phi\left(x_{t-1}, t-1\right).
\end{equation}}Compared to the SDS loss, ISM enhances the generation of 3D results by replacing the single-step DDPM with the multi-step DDIM, resulting in outputs with richer details. This approach effectively mitigates the issues of over-smoothness and blurriness in the results and notably accelerates the convergence rate. In this paper, we adopt ISM as the basis of our method to achieve more robust results.

Classifier-free guidance (CFG) ~\cite{ho2022classifier} is also employed in diffusion models with a guidance weight $\lambda_{cfg}$ to direct the unconditional score distribution to the conditional one. Specifically, $\delta(x_t,x_{t-1},t,t-1)$ with CFG is expressed as:
\revise{
\begin{equation}
\begin{aligned}
    \delta(x_t,x_{t-1};t,t-1)=&\ \epsilon_\phi\left(x_t; t\right)-\epsilon_\phi\left(x_{t-1}; t-1\right)\\
    +&\ \lambda_{cfg} \left(\epsilon_\phi\left(x_t; t, y\right)-\epsilon_\phi\left(x_t; t\right)\right).
\end{aligned}
\end{equation}
}
Inspired by CFG, we employ a similar formulation in our proposed method to direct the unstylized score distribution to a stylized one, thereby achieving stylization.

\subsection{Style-guided Texture Generation Pipeline}
\label{sec: generation}
Our stylized texture generation pipeline is depicted in Fig.~\ref{fig:pipeline}. 
The input encompasses an untextured 3D mesh denoted as $\mathcal{M}$, a reference image $I_{ref}$ providing the style. We utilize GPT-4 to extract a text prompt $y$ from the reference image $I_{ref}$, which characterizes the desired style and content, and a text prompt $y_{ref}$ that describes the content of the reference image.
%
%
Instead of directly optimizing the texture map in 2D space, we optimize a neural color field $\Gamma_\theta(p) = c$, where $p\in \mathcal{R}^3$ is the surface position of the 3D mesh and $c\in\mathcal{R}^3$ denotes the color. We represent the neural field by the hash-grid proposed by ~\cite{muller2022instant}. After optimization, the texture map can be sampled from the neural field, which is detailed in the Appendix ~\ref{sec: Texture Map Extraction}.

%
%
%
At each iteration, in addition to rendering the image $x_0$, we render the depth and normal maps indicating the geometric information, which are incorporated into the optimization by a geometry-aware ControlNet~\cite{zhang2023adding} to achieve geometry consistency. 
%
Besides, inspired by ISM~\cite{EnVision2023luciddreamer}, we incorporate a high-quality noise estimation method of ControlNet. Instead of simply sampling from a Gaussian distribution, we generate the noised $x_t$ by utilizing DDIM inversion to achieve superior noise estimation.

\begin{figure*}[t]
  \includegraphics[width=0.95\textwidth]{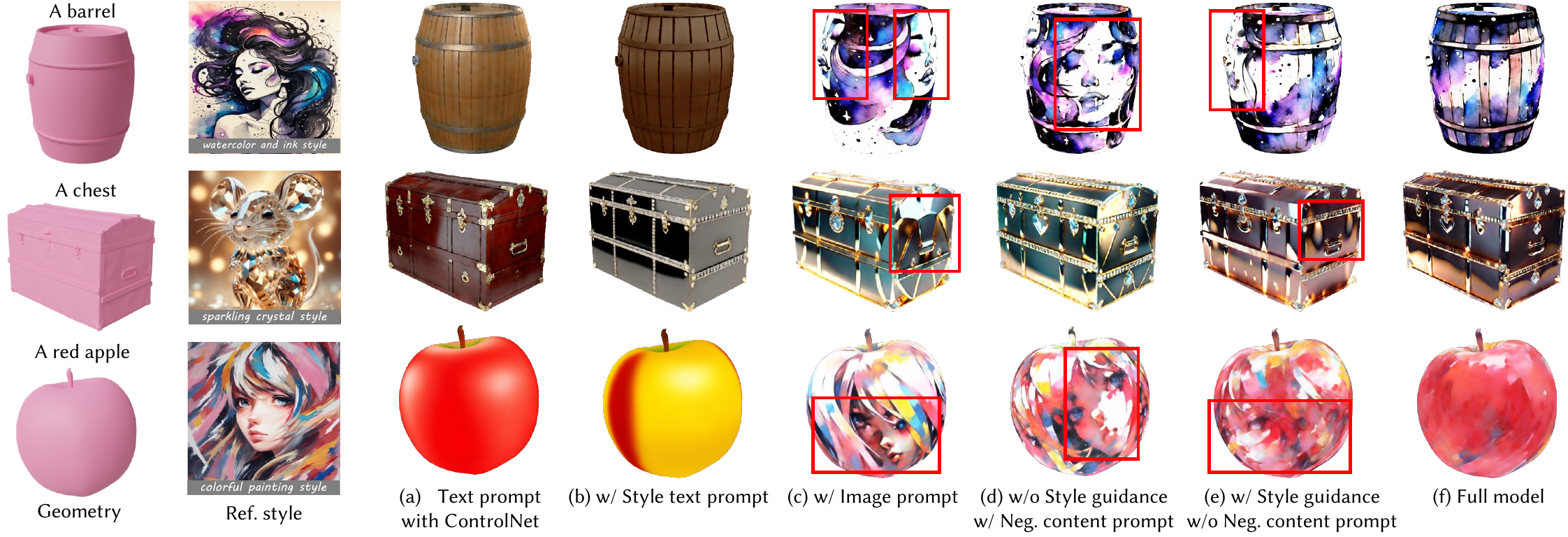}
  \caption{\textbf{Ablation study on style guidance}. (a) Baseline for text-to-texture. (b) Use ``in xxx style'' text prompts for style guidance. (c) Add the whole image prompt as guidance. (d) Add our style guidance strategy. 
  (e) Add content embedding of the reference image as a negative prompt. 
  (f) Full model with the style guidance strategy and content embedding of the reference image as a negative prompt.}
  \label{fig:ablation_all}
  \vspace{-3mm}
\end{figure*}
Specifically, the parameters $\theta$ of the neural field $\Gamma_\theta$ are optimized by our novel style-guided loss. The gradient of our loss is:
\begin{equation}
    \nabla_\theta \mathcal{L}_{\mathrm{ISM}}^\mathrm{style}(\theta)=\mathbb{E}_{t, c}\left[\omega(t)\delta(x_{t},x_{t-1};y,I_{ref},y_{ref},t,t-1) \frac{\partial g(\theta, c)}{\partial \theta}\right].
\end{equation}
The gradient updating direction $\delta(x_{t},x_{t-1};y,I_{ref},y_{ref},t,t-1)$ incorporates the style and content from the reference image $I_{ref}$ as well as two text prompts $y$ and $y_{ref}$. It is formulated as: 
\revise{
\begin{equation}
    \begin{aligned}
    \label{eq: final}
    &\delta(x_{t},x_{t-1};y,I_{ref},y_{ref},t,t-1)\\
       & =\  \epsilon_\phi(x_t;t)-\epsilon_\phi(x_{t-1};t-1)  \\
       & +\  \lambda_{cfg}(\epsilon_\phi(x_t;t,y)-\epsilon_\phi(x_t;t,y_{ref})) \\
       & +\  \delta_{style}(x_{t};y,I_{ref},y_{ref},t),
    \end{aligned}
\end{equation}
where $y_{ref}$ indicates the unintended content information. We integrate $y_{ref}$ into the CFG term $\epsilon_\phi(x_t;t,y)-\epsilon_\phi(x_t;t,y_{ref})$ as a negative prompt to reduce the content leakage artifacts.}
Besides, we explicitly employ a novel style guidance $\delta_{style}(x_{t};y,I_{ref},y_{ref},t)$ to direct the style of the rendered image to the desired one. 
The style guidance aims to reduce the score distribution divergence between the rendered images and the images with the desired style. Inspired by the classifier guidance, our style guidance can be formulated as:
\revise{
\begin{equation}
    \begin{aligned}
    \label{eq: style_final}
    &\delta_{style}(x_{t};y,I_{ref},y_{ref},t)\\
       & =\  \lambda_{style} (\epsilon_{style}(x_t;t,y,I_{ref},y_{ref})-\epsilon_\phi(x_t;t)),
    \end{aligned} \end{equation}}where $\lambda_{style}$ is a weight factor, and $\epsilon_{style}(x_t;t,y,I_{ref},y_{ref})$ predicts the distribution of the required style images.

\subsection{Style Score Distribution}
\label{sec: ism}
To achieve the style distribution for $\epsilon_{style}$, a possible way is to train a style-conditioned diffusion model, but it is time-consuming. Instead, inspired by~\cite{ye2023ip-adapter,wang2024instantstyle}, we shift the original non-style distribution of a pre-trained diffusion model to the desired one by injecting information from the reference image into the diffusion model.
Therefore, the core requirement is to extract style information from the reference image while disregarding content information.

Existing 2D style image generation studies~\cite{ye2023ip-adapter, wang2024instantstyle} have explored that the cross-attention mechanism in different transformer layers of a diffusion model exerts different effects on the content and style. Therefore, the stylized result can be achieved by injecting the features of the reference image into the layers that are responsible for style effects. However, a transformer layer can be in charge of both style and content because of the ambiguity in them. 
Leveraging such a layer to inject the reference image feature may introduce unintended content, while ignoring it may result in inaccuracies in style expressiveness, such as color tone shifting.
To address this, we aim to incorporate as many layers that are responsible for style effects as possible to maintain style expressiveness. The appendix contains detailed information about our leveraged transformer layers.
%
Simultaneously, to mitigate the influence of content from adding these layers, we propose explicitly disentangling the style and content from the image feature to extract a cleaner style.

To disentangle the content and style, we leverage the text content prompt $y_{ref}$ as the content guidance. 
Specifically, based on the multi-modal applications of the CLIP space, we encode the reference image and the text content prompt into the same space using a CLIP image encoder and a CLIP text encoder, respectively, resulting in image embedding and content embedding. 
While the content embedding encodes the majority of the content information of the image, text-based descriptions cannot align accurately with the abundant image information. 
Therefore, simply driving the image embedding towards the opposite of the content embedding direction cannot eliminate the content correctly. Driving too little does not influence the image content, while driving too much may alter the reference image's color tone.
To this end, we propose to decompose the image embedding into two components, with one component aligning with the content embedding explicitly. 
Specifically, we employ an orthogonal decomposition for content removal (ODCR):
\begin{equation}
\begin{aligned}
   & f_{g}^{ref}=E_{CLIP}^{img}(I_{ref}),  \quad f_{c}^{ref}=E_{CLIP}^{text}(y_{ref}), \\
   & f_{s}^{ref}=f_{g}^{ref}- \frac{f_{c}^{ref}({f_{g}^{ref}})^T f_{c}^{ref}}{||f_{c}^{ref}||^2_2}, 
\end{aligned}
\end{equation}
where $f_g^{ref}$ is the reference image embedding extracted by the CLIP's image encoder $E_{CLIP}^{img}$, and $f_c^{ref}$ is the content embedding extracted by the CLIP's text encoder $E_{CLIP}^{text}$.
After ODCR, we remain only the $f_{s}^{ref}$ to guide the diffusion model. The experiments in Sec.~\ref{sec:exp_style_guidance} demonstrate the superiority of our decomposition.



\section{Experiments and Results}
\subsection{Ablation Study}
We first conduct an ablation study to show the style effectiveness of each component of our method, including using $y_{ref}$ as the negative prompt and using our style guidance $\delta_{style}$ for disentangling and injecting the style. Then we dive into our style guidance to validate the effectiveness of our chosen transformer layers and the image embedding decomposition. Next, we validate that the geometry-aware ControlNet is beneficial to 3D consistency. Lastly, we conduct an experiment to show that using ISM achieves higher-quality results.

\subsubsection{Style effectiveness for each component}
\label{sec:exp_style_guidance}
As a baseline, we use a non-style text-to-texture generation that uses an ISM-based framework with a geometry-aware ControlNet to produce three outcomes, with ``a red apple'', ``a chest'', and \revise{``a barrel''} as the textual conditions, respectively. 
The results shown in Fig.~\ref{fig:ablation_all} (a) present multi-view consistency while not presenting any specific style. 
Then in Fig.~\ref{fig:ablation_all} (b), we add textual descriptions of the desired style in the prompt and hence these prompts become ``a barrel in watercolor and ink style'', ``a chest in sparkling crystal style'', and ``a red apple in a colorful painting style''. Although the results in  Fig.~\ref{fig:ablation_all} (b) demonstrate some color changes compared to the baseline, they fail to convey the style effectively. 
Image-based texture generation methods, such as ~\cite{ye2023ip-adapter}, take the reference image as the input and can achieve vivid style. However, as shown in Fig.~\ref{fig:ablation_all} (c), without disentangling the content and style, the content information of the reference image is incorrectly retained in the results.
We then showcase two variants of our method, one removes our style guidance (Fig.~\ref{fig:ablation_all} (d)) and another removes the negative prompt of the CFG term (Fig.~\ref{fig:ablation_all} (e)). Both methods achieve a vivid style and alleviate the content leakage artifacts. 
Lastly, our method shown in Fig.~\ref{fig:ablation_all} (f) exhibits high-quality results with vivid style while not presenting artifacts.

\begin{figure}[t]
  \includegraphics[width=0.99\linewidth]{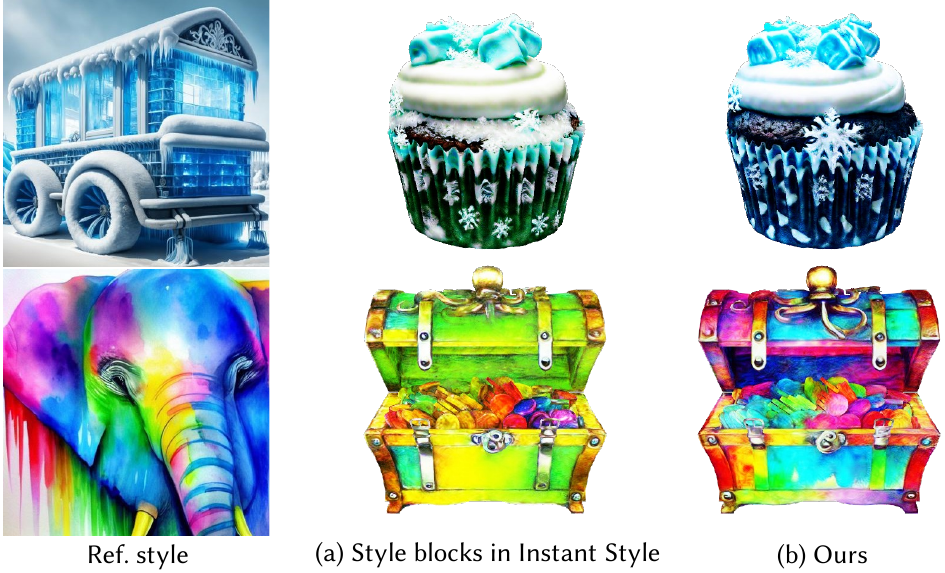}
  \caption{\revise{Stylized texture results obtained using various transformer layer style injection strategies.  
  } 
  The Prompts are ``a cupcake in ice and snow covered style'' and ``a wooden treasure chest with metal accents and locks in colorful drawing style''.}
  \label{fig:blocks}
  \vspace{-3mm}
\end{figure}

\begin{figure}[t]
  \includegraphics[width=0.99\linewidth]{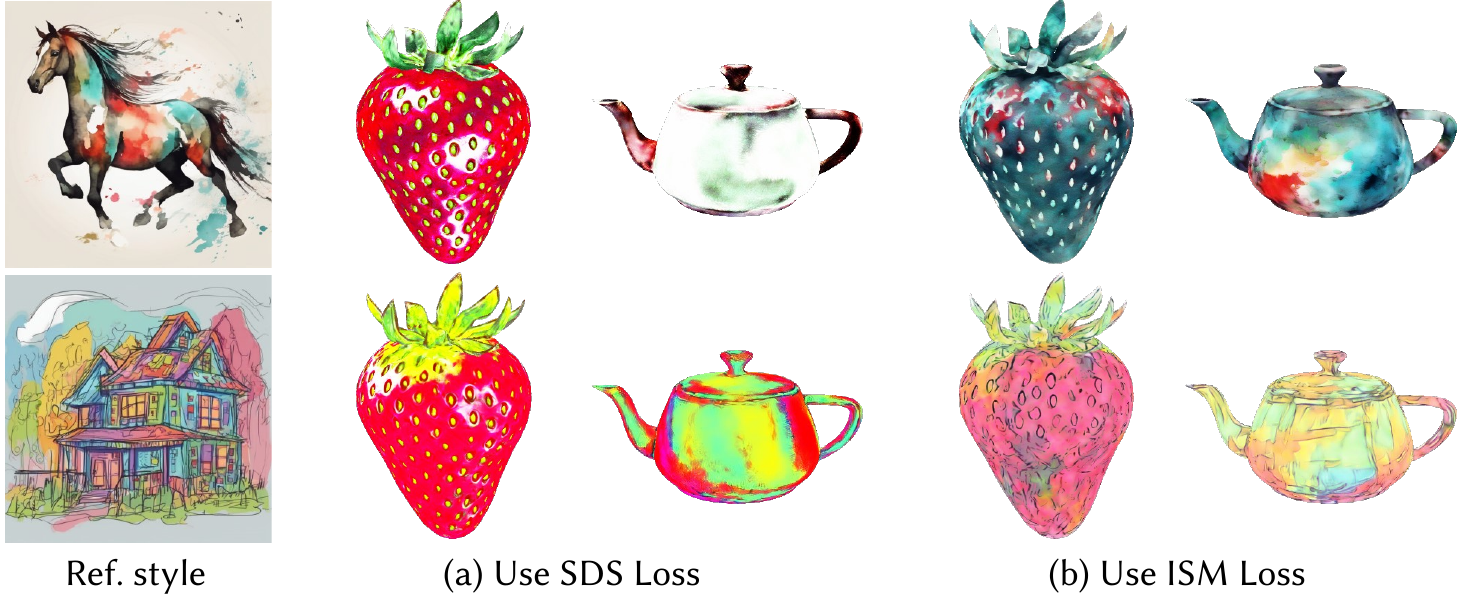}
  \caption{Results using our style-content decoupling method with SDS loss (a) and ISM loss (b) for the prompts ``a strawberry/teapot in colorful graffiti style'' and ``a strawberry/teapot in Chinese ink paint style''.}
  \label{fig:SDS_ISM}
  \vspace{-6mm}
\end{figure}

\subsubsection{Style guidance}
\begin{figure}[t]
  \includegraphics[width=0.99\linewidth]{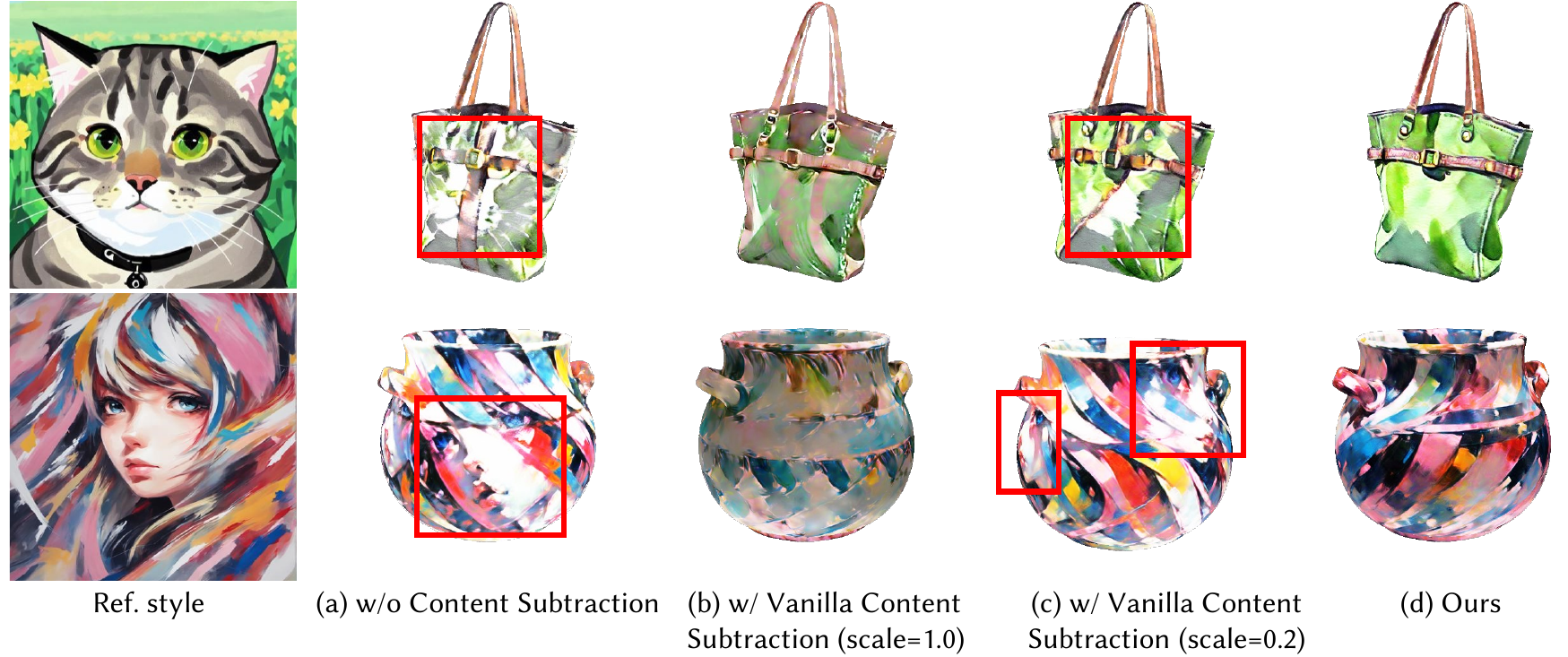}
  \caption{Stylized texture results achieved using different content removal strategies in CLIP space. 
  The prompts are ``a hand bag in watercolor sketch style'' and ``a pot in a colorful painting style''.}
  \label{fig:ODCIE}
  \vspace{-3mm}
\end{figure}

Our style guidance $\delta_{style}$ is carefully designed to preserve style expressiveness while not leading to content leakage by two aspects. 
Firstly, in terms of style injection in transformer layers, unlike existing 2D style image generation methods that do not consider transformer layers that are more responsible for content than style, we use all transformer layers that impact style to achieve style consistency in multiple views. 
As shown in Fig. \ref{fig:blocks}, incorporating only the transformer layers used by 2D style image generation methods can result in a color tone that deviates from that of the reference image.
Secondly, we explicitly decompose the reference image embedding within the CLIP space to disentangle the style and content. As shown in Fig.~\ref{fig:ODCIE} (a), incorporating the complete image embedding into the diffusion model leads to severe content leakage artifacts. 
Besides, without disentangling the style and content, driving the image embedding by the content embedding easily results in artifacts. 
For instance, greatly altering the image embedding can lead to inaccurate color expressiveness (Fig. \ref{fig:ODCIE} (b)), while slight modifications cause content leakage (Fig. \ref{fig:ODCIE} (c)).
In Fig. \ref{fig:ODCIE} (d), our method presents a superior performance in both style expressiveness and content removal.

\subsubsection{ISM vs SDS}

To achieve superior quality, we utilize an ISM-based optimization framework instead of SDS~\cite{poole2022dreamfusion}.
As illustrated in Fig.~\ref{fig:SDS_ISM}, replacing our ISM loss with the SDS loss exhibits over-saturation and over-smoothness and severely undermines the style expressiveness.

\subsubsection{Geometry-aware ControlNet}
\begin{figure}[t]
  \includegraphics[width=0.99\linewidth]{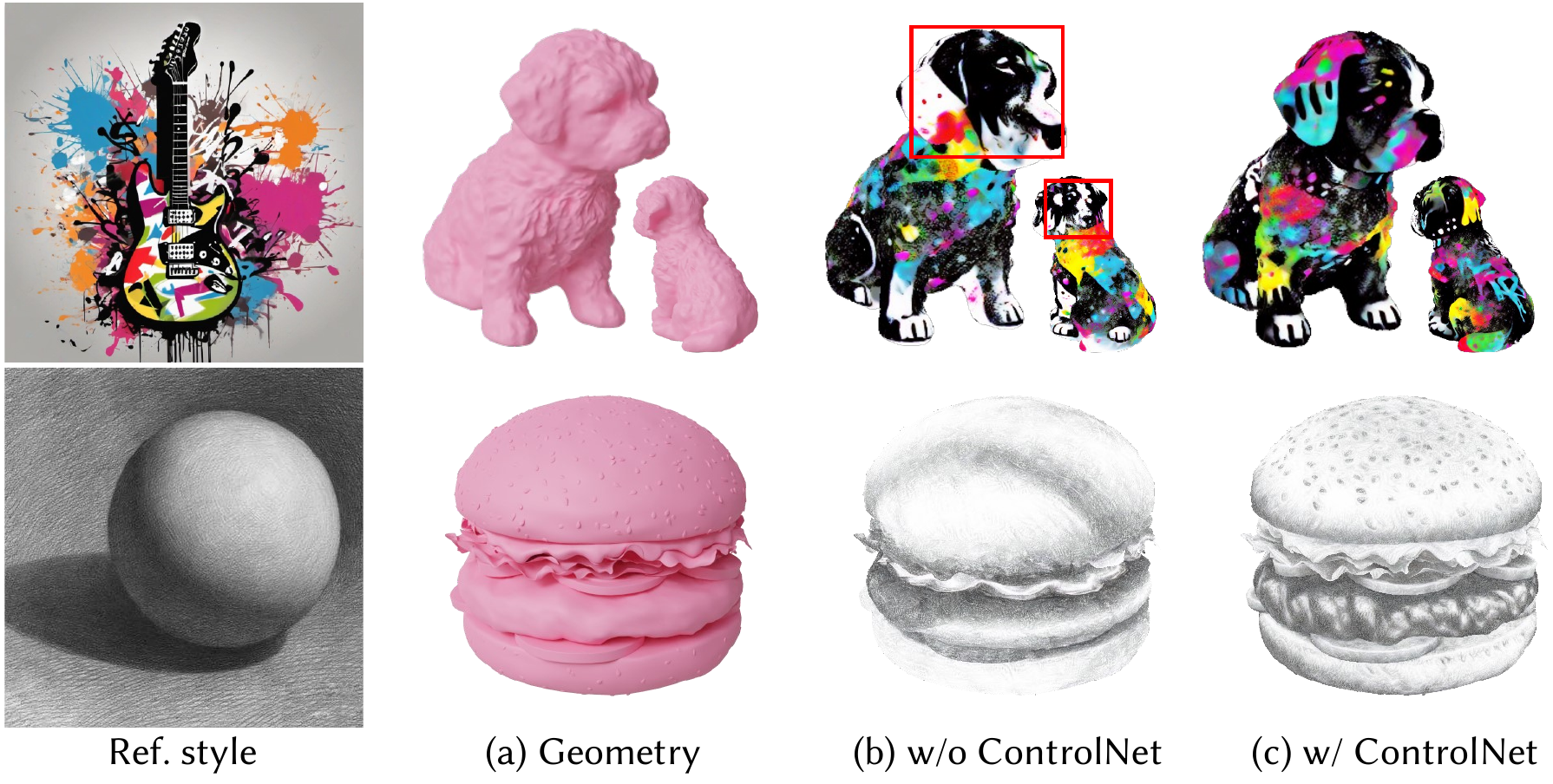}
  \caption{\textbf{Ablation study on geometric ControlNet.} The prompts are ``a dog in graffiti style'' and ``a hamburger in sketch style''.} 
  \label{fig:controlnet}
  \vspace{-6mm}
\end{figure}

Our method uses a geometry-aware ControlNet that receives the rendered depth and normal map as inputs. To validate its effectiveness in preserving 3D consistency and geometrical details, we conduct an experiment using a vanilla diffusion model.
As shown in Fig.~\ref{fig:controlnet}, the geometry-aware ControlNet greatly enhances the detail of textures, particularly in models with complex geometries (e.g., the hamburger). Furthermore, it also aids in eliminating the Janus problem as shown in the first row of Fig.~\ref{fig:controlnet}.

\begin{figure*}[t]
  \includegraphics[width=0.99\linewidth]{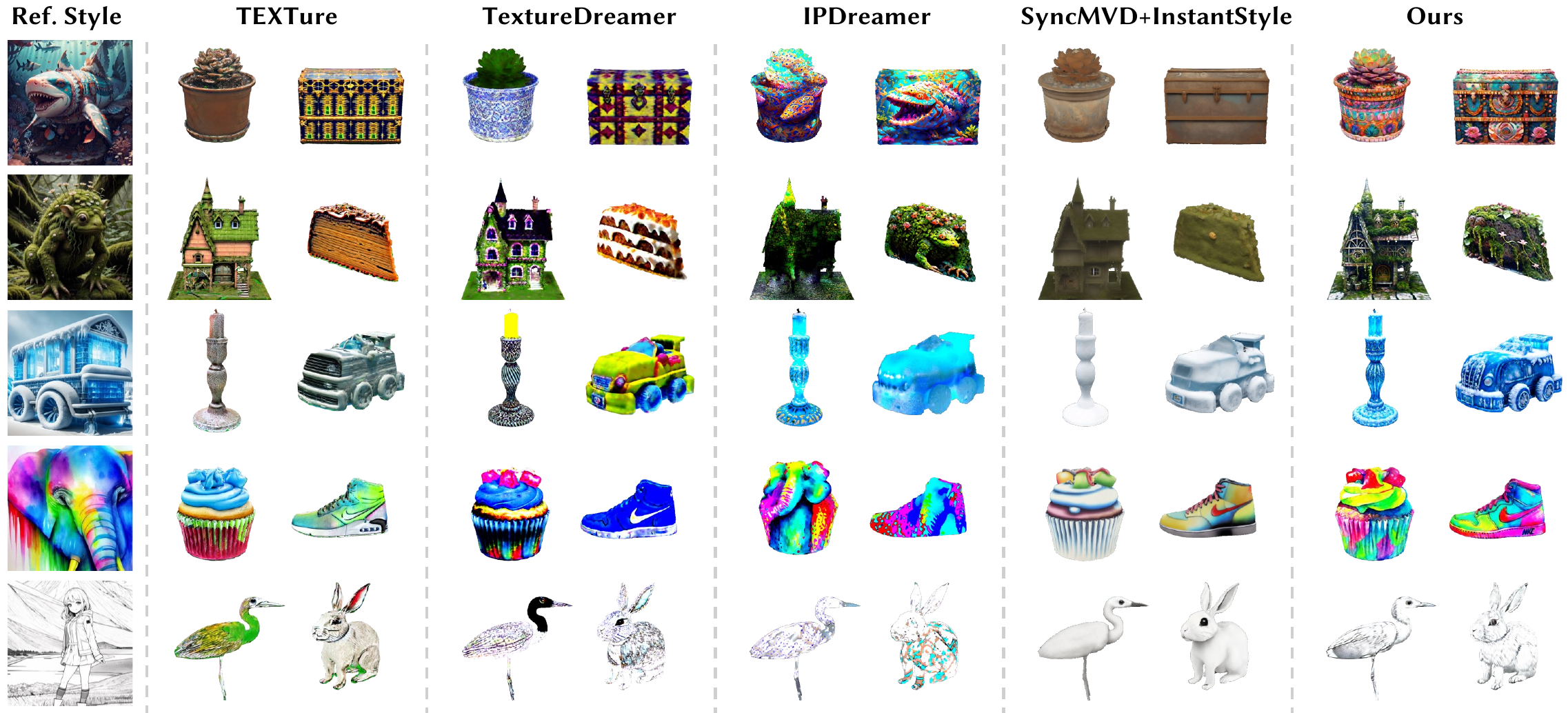}
  \caption{Qualitative comparison to TEXTure~\cite{richardson2023texture}, TextureDreamer~\cite{yeh2024texturedreamer}, IPDreamer~\cite{zeng2023ipdreamer}, and SyncMVD~\cite{liu2023text}.}
  \label{fig:comparison_paper}
  \vspace{-2mm}
\end{figure*}

\subsection{Comparison}
We compare our method to several state-of-the-art methods for image-guided 3D generation, namely TEXTure~\cite{richardson2023texture}, TextureDreamer~\cite{yeh2024texturedreamer}, IPDreamer~\cite{zeng2023ipdreamer}, and a text-based texture generation method SyncMVD~\cite{liu2023text}. Since IPDreamer~\cite{zeng2023ipdreamer} synthesizes 3D geometry in addition to texture, we fix the geometry and concentrate solely on texture synthesis. Besides, SyncMVD~\cite{liu2023text} synthesizes 2D images across multiple views rather than using score distillation sampling and hence cannot incorporate our style guidance. For a fair comparison, we incorporate a 2D image-guided generation method~\cite{wang2024instantstyle} into SyncMVD.  

\begin{table}
\caption{\textbf{User study results.} Participants are asked to evaluate the overall quality, style fidelity, and content removal of the generated results by giving scores ($\in [1,5]$) to the rendering videos. This table shows the average scores given by 37 participants.}
\vspace{-3mm}
\centering
\scalebox{0.95}{
\setlength{\tabcolsep}{2pt}
\begin{tabular}{lcccccc}
\toprule

Method & 
\begin{tabular}[c]
{@{}c@{}}Overall Quality$\uparrow$ \end{tabular}  &
\begin{tabular}[c]
{@{}c@{}}Style Fidelity$\uparrow$ \end{tabular} & 
\begin{tabular}[c]
{@{}c@{}}Content Removal$\uparrow$ \end{tabular} \\

\midrule
TEXTure      & 2.62 & 2.01 &  2.83 \\
TextureDreamer        & 2.29 &  1.93  & 2.68 \\
IPDreamer     & 3.02 & 3.45 &  3.01\\
SyncMVD    & 3.07& 2.67 & 3.23  \\
Ours    & \textbf{4.60} & \textbf{4.61} & \textbf{4.36} \\
\bottomrule
\end{tabular}
}

\label{tab:user_study}
\vspace{-6mm}
\end{table}

\subsubsection{Qualitative Comparison}
\begin{figure*}
  \includegraphics[width=\textwidth]{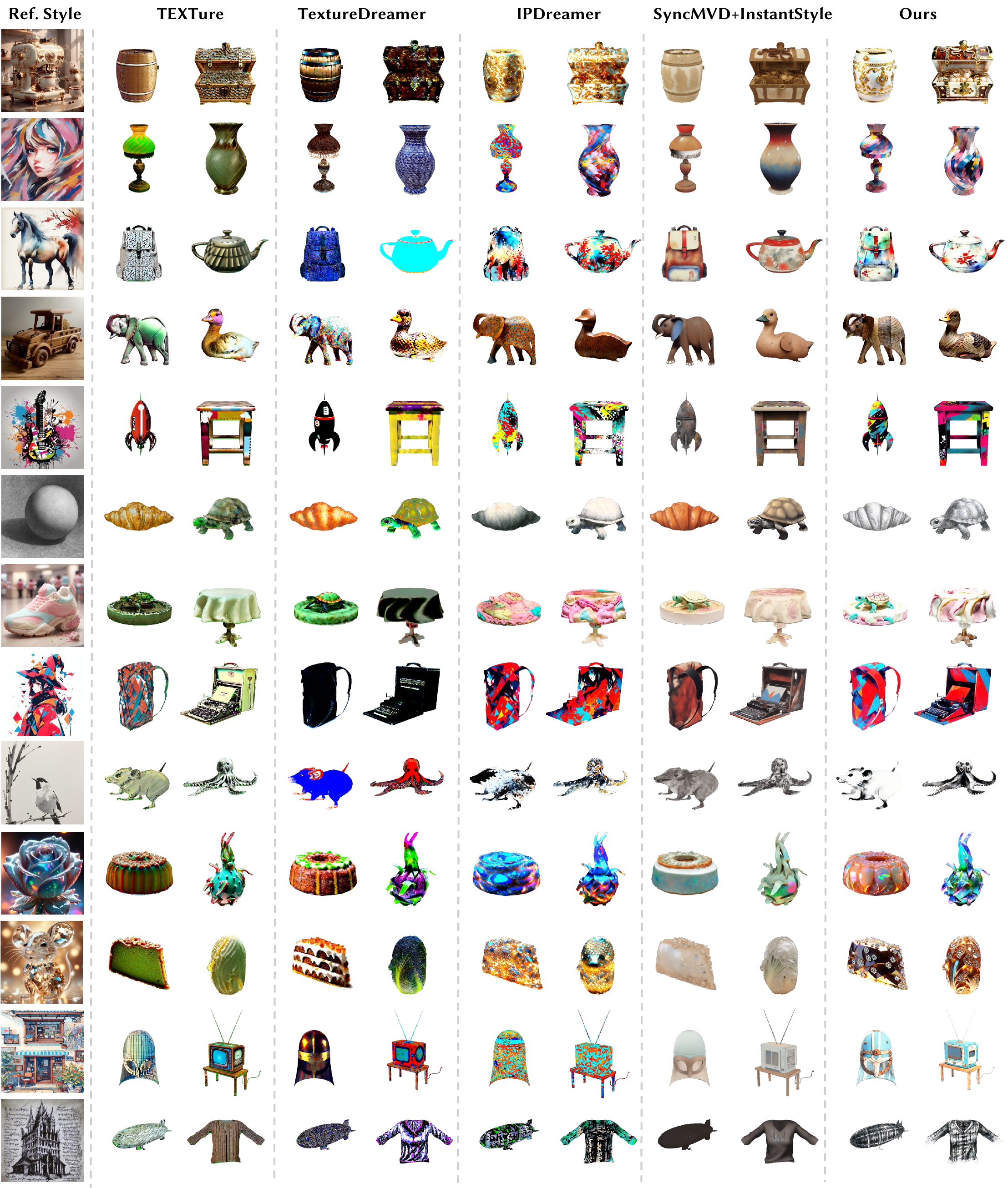}
  \caption{More qualitative comparison with TEXTure~\cite{richardson2023texture}, TextureDreamer~\cite{yeh2024texturedreamer}, IPDreamer~\cite{zeng2023ipdreamer}, and SyncMVD~\cite{liu2023text}.}
  \label{fig:comparison}
\end{figure*}

Fig. \ref{fig:comparison_paper} and Fig.~\ref{fig:comparison} provide qualitative comparisons between the baseline methods and our proposed approach. Both TEXTure and TextureDreamer utilize reference images to fine-tune the diffusion model, with the generated texture heavily relying on the performance of fine-tuning. However, given only a single reference image, the fine-tuned diffusion model either overfits or fails to accurately extract the image style, leading to incorrect results when applied to a mesh whose subject does not match the reference image. IPDreamer does not separate the style and content of the reference image during generation, resulting in a significant content leakage issue. Additionally, the usage of the SDS leads to over-saturation. While SyncMVD can synthesize multi-view images that exhibit some extent of the style, it suffers from balancing between the multi-view consistency, the image guidance and the classifier term, leading to overly smooth, detail-lacking, and style drifting results.
In contrast, our results demonstrate superior performance in terms of detail representation and style fidelity compared to all other methods.

\begin{table}
\caption{\textbf{Quantitative comparison results.} We utilize the Gram Matrix Distance to measure style fidelity, and use the CLIP score to measure the semantic alignment between the prompts and the results.}
\vspace{-3mm}
\centering
\scalebox{1.0}{
\setlength{\tabcolsep}{2pt}
\begin{tabular}{lcccccc}
\toprule

Method & 
\begin{tabular}[c]
{@{}c@{}}Gram Matrix Distance$\downarrow$ \end{tabular} & 
\begin{tabular}[c]{@{}c@{}}CLIP Score$\uparrow$ \end{tabular}  \\


\midrule
TEXTure      & \revise{0.830} &   \revise{68.01} \\
TextureDreamer        &   \revise{0.947}  & \revise{68.57} \\
IPDreamer     & \revise{0.910} &   \revise{61.81}\\
SyncMVD    & \revise{0.920}&  \revise{69.60}  \\
Ours    & \revise{\textbf{0.723}} &  \revise{\textbf{73.66}} \\
\bottomrule

\end{tabular}
}
\vspace{-6mm}
\label{tab:quantitative}
\end{table}

\subsubsection{Quantitative Comparison}

We first conduct user study using 12 styles and 24 meshes to evaluate the results of all methods regarding quality, style fidelity, and content removal. For each style, we use each method to generate textures for 2 meshes, respectively.
We ask 37 participants to assign a score range from 1 to 5 to the synthesized results of all methods. The higher score indicates the better performance. The results are shown in Tab.~\ref{tab:user_study}. Among all methods, our method achieves the superior performance in terms of all metrics.

In addition to the user study, we use the common metrics for image generation methods to evaluate all methods in terms of style fidelity and semantic alignment. \revise{
For 25 randomly chosen styles, we use each style to generate stylized textures for 4 unique, randomly selected meshes from Objaverse~\cite{deitke2023objaverse}, totaling 100 different results. We then render four views per result to compute the metrics.} The style's fidelity is measured by 
the Gram metrics difference~\cite{johnson2016perceptual} between the rendered images and the reference images. Besides, the semantic alignment between the prompts and the rendered image is measured by the CLIP Score~\cite{hessel2021clipscore}.
As shown in Tab.~\ref{tab:quantitative}, our method outperforms all other methods in achieving the best style fidelity and text alignment. The details of these metrics are outlined in the Appendix ~\ref{sec: Quantitative Evaluation Matrix}.

\subsection{Results}
\begin{figure*}
  \includegraphics[width=\textwidth]{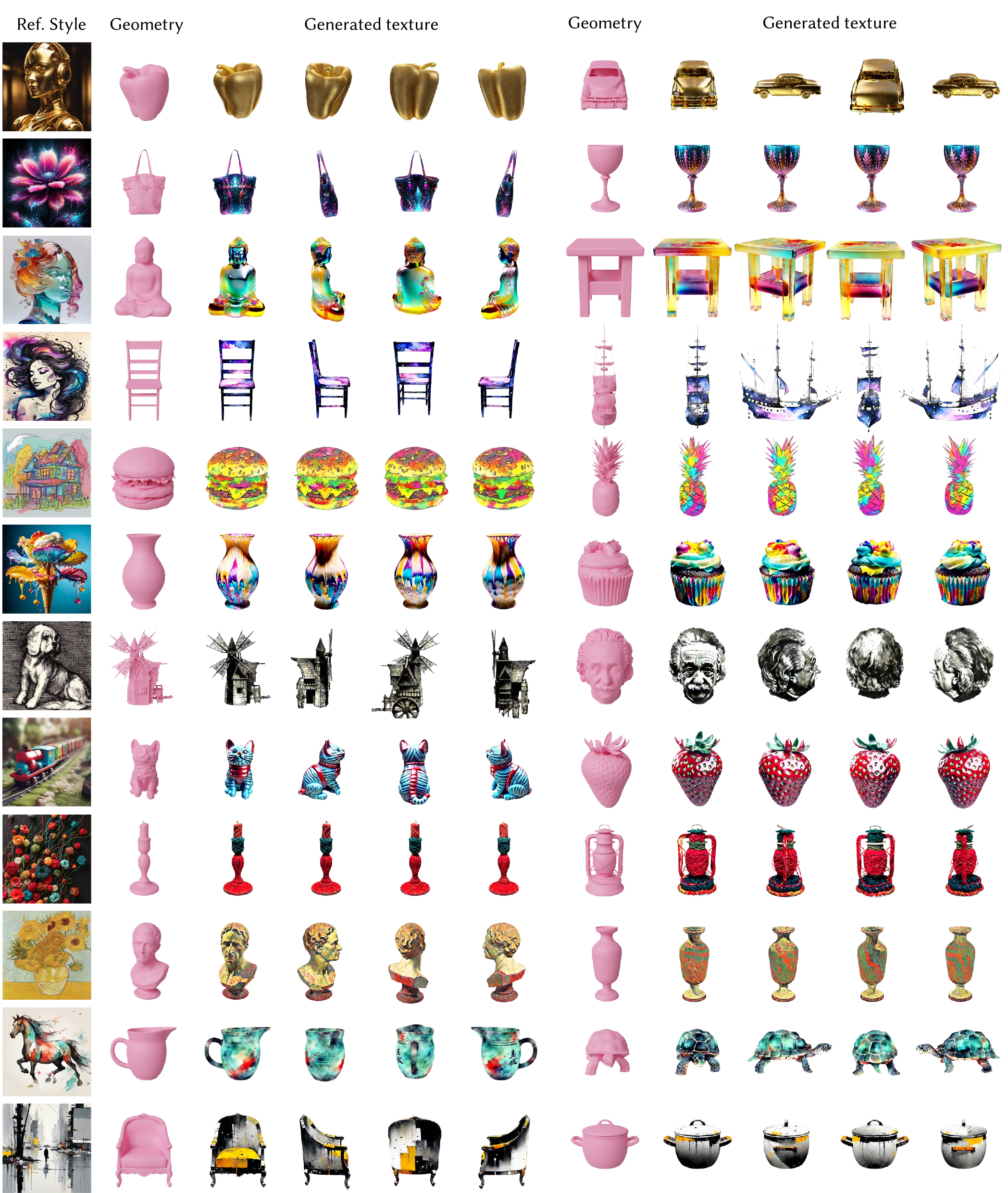}
  \caption{Results of StyleTex. For each style, we generate textures for two meshes and showcase four different rendered views.}
  \label{fig:results1}
\end{figure*}

An NVIDIA RTX 4090 GPU is used for the optimization process, which takes about 15 minutes to synthesize a texture map for each mesh.
We demonstrate the robustness of our method using a diverse range of reference images, including various artistic styles such as ``sketching'' and ``ink wash painting'', different materials like ``gold'' and ``wool'', as well as various patterns and brush strokes. The generated results shown in Fig.~\ref{fig:results1} maintain multi-view consistency, align with the geometric details of the models, and adhere to the style of the reference image.

In addition, we demonstrate that our method can be practically used for games or films, which requires generating consistent style for all meshes. As shown in Fig.~\ref{fig:teaser}, we create textures for various objects that share the same style given a reference image, resulting in scenes that are harmonious and aesthetically pleasing.




\section{Limitations and Conclusions}
\subsection{Limitations} 

\begin{figure}[t]
  \includegraphics[width=0.99\linewidth]{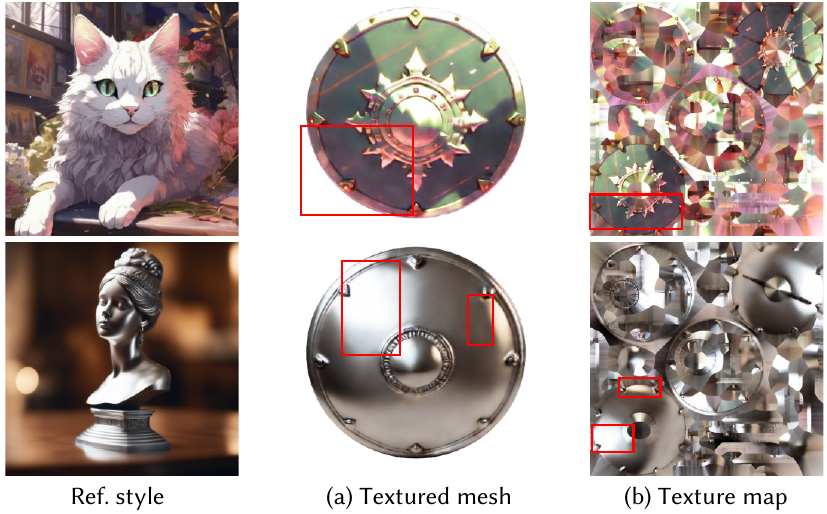}
  \vspace{-3mm}
  \caption{\revise{
  Artifacts caused by baked-in highlights or shadows in generated textures (b). The red boxes represent unintended baked-in shadows (a,b) (upper) and highlights (a,b) (bottom). 
  }}
  \label{fig:limitation}
  \vspace{-4mm}
\end{figure}

Despite the successful generation of high-quality textures that align with the style of the reference image, our method presents several limitations. 
%
\revise{
To begin, unlike PBR materials generation methods ~\cite{zhang2024clay, zhang2024dreammat}, the influence of style prevents us from identifying a universally applicable rendering model, making it difficult to define and decouple the highlights and shadows contained in textures. 
This issue may result in baked-in highlights
or shadows in the generated textures, as shown in Fig. \ref{fig:limitation}. 
Second, our method's distillation time is relatively long, which limits its use in an interactive environment.}
%
Future work could potentially accelerate our method by integrating recent advancements in diffusion models and representations. 
Finally, as style is the result of a combination of various elements (including material, brush strokes, tone, and painting style), our method is unable to extract or adjust any of these elements individually.

\subsection{Conclusions} This paper presents StyleTex, a novel stylized texture generation approach for the given mesh, guided by a single reference image and text prompts. StyleTex leverages an ISM-based generative framework, incorporating both style guidance and geometric control. The key advantage of our method is a novel strategy for disentangling style and content information, which effectively addresses the prevalent issues of content leakage and style drift in 3D stylized textures. 
By utilizing a single stylized image as the reference, StyleTex can generate textures that exhibit similar styles, thereby enabling the automatic creation of visually compelling and immersive virtual environments for games or films.

\begin{acks}
Xiaogang Jin was supported by Key R\&D Program of Zhejiang (No. 2024C01069).
\end{acks}

\revise{
\begin{figure*}
  \includegraphics[width=\textwidth]{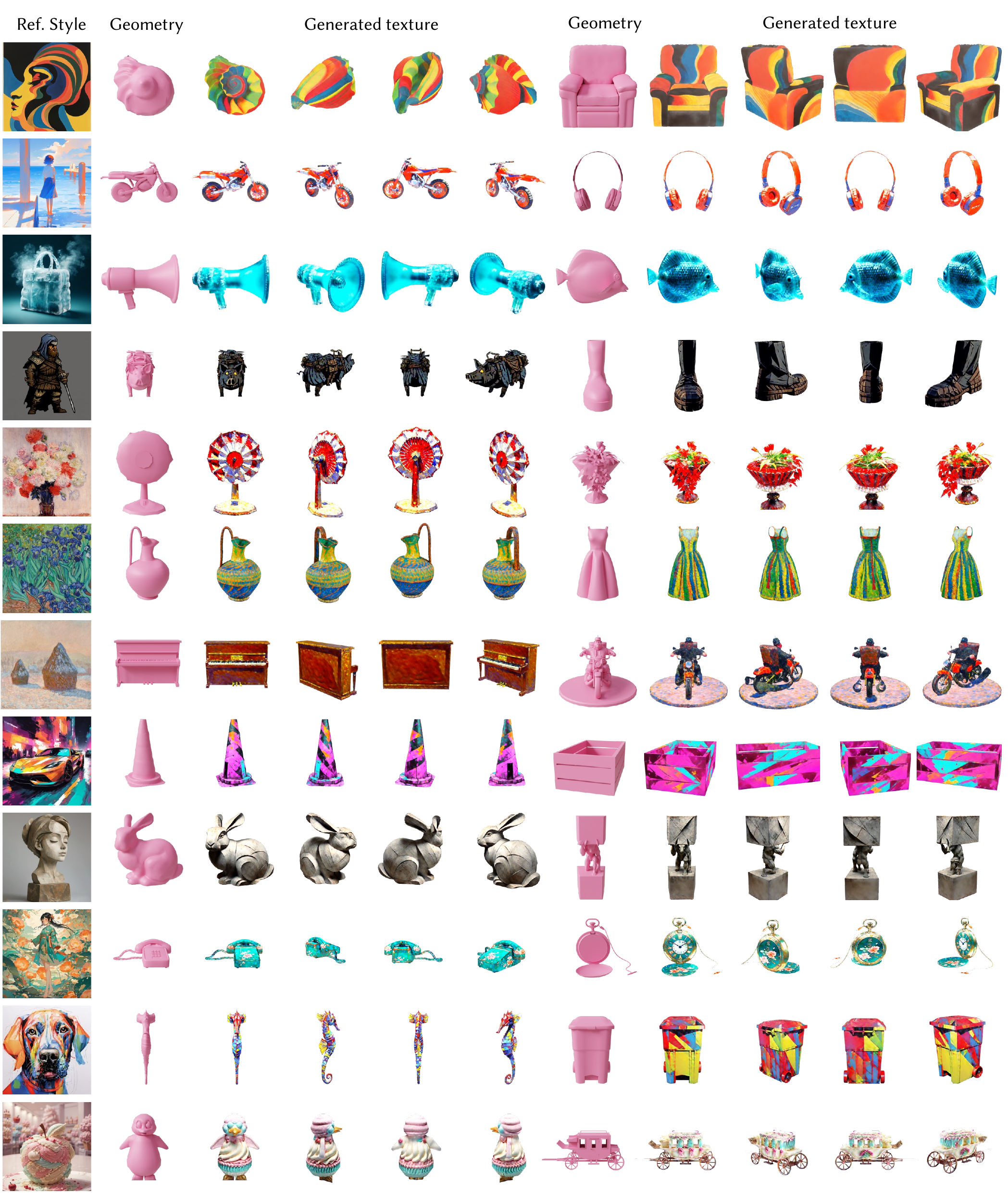}
  \caption{\revise{Results of StyleTex. For each style, we generate textures for two meshes and showcase four different rendered views.}}
  \label{fig:results2}
\end{figure*}
}

\bibliographystyle{ACM-Reference-Format}
\bibliography{ref}


\begin{thebibliography}{80}


\ifx \showCODEN    \undefined \def \showCODEN     #1{\unskip}     \fi
\ifx \showDOI      \undefined \def \showDOI       #1{#1}\fi
\ifx \showISBNx    \undefined \def \showISBNx     #1{\unskip}     \fi
\ifx \showISBNxiii \undefined \def \showISBNxiii  #1{\unskip}     \fi
\ifx \showISSN     \undefined \def \showISSN      #1{\unskip}     \fi
\ifx \showLCCN     \undefined \def \showLCCN      #1{\unskip}     \fi
\ifx \shownote     \undefined \def \shownote      #1{#1}          \fi
\ifx \showarticletitle \undefined \def \showarticletitle #1{#1}   \fi
\ifx \showURL      \undefined \def \showURL       {\relax}        \fi
\providecommand\bibfield[2]{#2}
\providecommand\bibinfo[2]{#2}
\providecommand\natexlab[1]{#1}
\providecommand\showeprint[2][]{arXiv:#2}

\bibitem[Achiam et~al\mbox{.}(2023)]%
        {achiam2023gpt}
\bibfield{author}{\bibinfo{person}{Josh Achiam}, \bibinfo{person}{Steven Adler}, \bibinfo{person}{Sandhini Agarwal}, \bibinfo{person}{Lama Ahmad}, \bibinfo{person}{Ilge Akkaya}, \bibinfo{person}{Florencia~Leoni Aleman}, \bibinfo{person}{Diogo Almeida}, \bibinfo{person}{Janko Altenschmidt}, \bibinfo{person}{Sam Altman}, \bibinfo{person}{Shyamal Anadkat}, {et~al\mbox{.}}} \bibinfo{year}{2023}\natexlab{}.
\newblock \showarticletitle{Gpt-4 technical report}.
\newblock \bibinfo{journal}{\emph{arXiv preprint arXiv:2303.08774}} (\bibinfo{year}{2023}).
\newblock


\bibitem[Agarwal et~al\mbox{.}(2023)]%
        {agarwal2023image}
\bibfield{author}{\bibinfo{person}{Aishwarya Agarwal}, \bibinfo{person}{Srikrishna Karanam}, \bibinfo{person}{Tripti Shukla}, {and} \bibinfo{person}{Balaji~Vasan Srinivasan}.} \bibinfo{year}{2023}\natexlab{}.
\newblock \showarticletitle{An Image Is Worth Multiple Words: Multi-Attribute Inversion for Constrained Text-to-Image Synthesis}.
\newblock \bibinfo{journal}{\emph{arXiv preprint arXiv:2311.11919}} (\bibinfo{year}{2023}).
\newblock


\bibitem[An et~al\mbox{.}(2021)]%
        {artflow2021}
\bibfield{author}{\bibinfo{person}{Jie An}, \bibinfo{person}{Siyu Huang}, \bibinfo{person}{Yibing Song}, \bibinfo{person}{Dejing Dou}, \bibinfo{person}{Wei Liu}, {and} \bibinfo{person}{Jiebo Luo}.} \bibinfo{year}{2021}\natexlab{}.
\newblock \showarticletitle{ArtFlow: Unbiased image style transfer via reversible neural flows}. In \bibinfo{booktitle}{\emph{{IEEE/CVF} Conference on Computer Vision and Pattern Recognition, {CVPR} 2021}}. \bibinfo{publisher}{{IEEE}}, \bibinfo{pages}{862--871}.
\newblock


\bibitem[Cao et~al\mbox{.}(2023)]%
        {cao2023texfusion}
\bibfield{author}{\bibinfo{person}{Tianshi Cao}, \bibinfo{person}{Karsten Kreis}, \bibinfo{person}{Sanja Fidler}, \bibinfo{person}{Nicholas Sharp}, {and} \bibinfo{person}{KangXue Yin}.} \bibinfo{year}{2023}\natexlab{}.
\newblock \showarticletitle{TexFusion: Synthesizing 3D Textures with Text-Guided Image Diffusion Models}. In \bibinfo{booktitle}{\emph{2023 {IEEE/CVF} International Conference on Computer Vision, {ICCV} 2023}}. \bibinfo{publisher}{{IEEE}}, \bibinfo{pages}{4146--4158}.
\newblock


\bibitem[Chen et~al\mbox{.}(2017)]%
        {chen2017stylebank}
\bibfield{author}{\bibinfo{person}{Dongdong Chen}, \bibinfo{person}{Lu Yuan}, \bibinfo{person}{Jing Liao}, \bibinfo{person}{Nenghai Yu}, {and} \bibinfo{person}{Gang Hua}.} \bibinfo{year}{2017}\natexlab{}.
\newblock \showarticletitle{StyleBank: An Explicit Representation for Neural Image Style Transfer}. In \bibinfo{booktitle}{\emph{2017 {IEEE} Conference on Computer Vision and Pattern Recognition, {CVPR} 2017}}. \bibinfo{pages}{2770--2779}.
\newblock


\bibitem[Chen et~al\mbox{.}(2023c)]%
        {chen2023text2tex}
\bibfield{author}{\bibinfo{person}{Dave~Zhenyu Chen}, \bibinfo{person}{Yawar Siddiqui}, \bibinfo{person}{Hsin-Ying Lee}, \bibinfo{person}{Sergey Tulyakov}, {and} \bibinfo{person}{Matthias Nie{\ss}ner}.} \bibinfo{year}{2023}\natexlab{c}.
\newblock \showarticletitle{Text2Tex: Text-driven Texture Synthesis via Diffusion Models}. In \bibinfo{booktitle}{\emph{2023 {IEEE/CVF} International Conference on Computer Vision, {ICCV} 2023}}. \bibinfo{publisher}{{IEEE}}, \bibinfo{pages}{18512--18522}.
\newblock


\bibitem[Chen et~al\mbox{.}(2023b)]%
        {chen2023controlstyle}
\bibfield{author}{\bibinfo{person}{Jingwen Chen}, \bibinfo{person}{Yingwei Pan}, \bibinfo{person}{Ting Yao}, {and} \bibinfo{person}{Tao Mei}.} \bibinfo{year}{2023}\natexlab{b}.
\newblock \showarticletitle{ControlStyle: Text-Driven Stylized Image Generation Using Diffusion Priors}.
\newblock \bibinfo{journal}{\emph{arXiv preprint arXiv:2311.05463}} (\bibinfo{year}{2023}).
\newblock


\bibitem[Chen et~al\mbox{.}(2023a)]%
        {Chen_2023_ICCV}
\bibfield{author}{\bibinfo{person}{Rui Chen}, \bibinfo{person}{Yongwei Chen}, \bibinfo{person}{Ningxin Jiao}, {and} \bibinfo{person}{Kui Jia}.} \bibinfo{year}{2023}\natexlab{a}.
\newblock \showarticletitle{Fantasia3D: Disentangling Geometry and Appearance for High-quality Text-to-3D Content Creation}. In \bibinfo{booktitle}{\emph{2023 {IEEE/CVF} International Conference on Computer Vision, {ICCV} 2023}}. \bibinfo{publisher}{{IEEE}}, \bibinfo{pages}{22189--22199}.
\newblock


\bibitem[Chen and Schmidt(2016)]%
        {Chen2016FastPS}
\bibfield{author}{\bibinfo{person}{Tian~Qi Chen} {and} \bibinfo{person}{Mark~W. Schmidt}.} \bibinfo{year}{2016}\natexlab{}.
\newblock \showarticletitle{Fast Patch-based Style Transfer of Arbitrary Style}.
\newblock \bibinfo{journal}{\emph{arXiv preprint arXiv:1612.04337}} (\bibinfo{year}{2016}).
\newblock


\bibitem[Chen et~al\mbox{.}(2022)]%
        {chen2022AUVNET}
\bibfield{author}{\bibinfo{person}{Zhiqin Chen}, \bibinfo{person}{Kangxue Yin}, {and} \bibinfo{person}{Sanja Fidler}.} \bibinfo{year}{2022}\natexlab{}.
\newblock \showarticletitle{AUV-Net: Learning Aligned UV Maps for Texture Transfer and Synthesis}. In \bibinfo{booktitle}{\emph{{IEEE/CVF} Conference on Computer Vision and Pattern Recognition, CVPR 2022}}. \bibinfo{publisher}{{IEEE}}, \bibinfo{pages}{1455--1464}.
\newblock


\bibitem[Cheskidova et~al\mbox{.}(2023)]%
        {Geometry_Aware_Texturing}
\bibfield{author}{\bibinfo{person}{Evgeniia Cheskidova}, \bibinfo{person}{Aleksandr Arganaidi}, \bibinfo{person}{Daniel-Ionut Rancea}, {and} \bibinfo{person}{Olaf Haag}.} \bibinfo{year}{2023}\natexlab{}.
\newblock \showarticletitle{Geometry Aware Texturing}. In \bibinfo{booktitle}{\emph{SIGGRAPH Asia 2023 Posters}} \emph{(\bibinfo{series}{SA '23})}. \bibinfo{publisher}{Association for Computing Machinery}, Article \bibinfo{articleno}{21}, \bibinfo{numpages}{2}~pages.
\newblock
\urldef\tempurl%
\url{https://doi.org/10.1145/3610542.3626152}
\showDOI{\tempurl}


\bibitem[Dai et~al\mbox{.}(2017)]%
        {dai2017scannet}
\bibfield{author}{\bibinfo{person}{Angela Dai}, \bibinfo{person}{Angel~X. Chang}, \bibinfo{person}{Manolis Savva}, \bibinfo{person}{Maciej Halber}, \bibinfo{person}{Thomas~A. Funkhouser}, {and} \bibinfo{person}{Matthias Nie{\ss}ner}.} \bibinfo{year}{2017}\natexlab{}.
\newblock \showarticletitle{ScanNet: Richly-Annotated 3D Reconstructions of Indoor Scenes}. In \bibinfo{booktitle}{\emph{2017 {IEEE} Conference on Computer Vision and Pattern Recognition, {CVPR} 2017}}. \bibinfo{publisher}{{IEEE} Computer Society}, \bibinfo{pages}{2432--2443}.
\newblock


\bibitem[Deitke et~al\mbox{.}(2023)]%
        {deitke2023objaverse}
\bibfield{author}{\bibinfo{person}{Matt Deitke}, \bibinfo{person}{Dustin Schwenk}, \bibinfo{person}{Jordi Salvador}, \bibinfo{person}{Luca Weihs}, \bibinfo{person}{Oscar Michel}, \bibinfo{person}{Eli VanderBilt}, \bibinfo{person}{Ludwig Schmidt}, \bibinfo{person}{Kiana Ehsani}, \bibinfo{person}{Aniruddha Kembhavi}, {and} \bibinfo{person}{Ali Farhadi}.} \bibinfo{year}{2023}\natexlab{}.
\newblock \showarticletitle{Objaverse: {A} Universe of Annotated 3D Objects}. In \bibinfo{booktitle}{\emph{{IEEE/CVF} Conference on Computer Vision and Pattern Recognition, {CVPR} 2023}}. \bibinfo{publisher}{{IEEE}}, \bibinfo{pages}{13142--13153}.
\newblock


\bibitem[Dumoulin et~al\mbox{.}(2017)]%
        {dumoulin2017alearned}
\bibfield{author}{\bibinfo{person}{Vincent Dumoulin}, \bibinfo{person}{Jonathon Shlens}, {and} \bibinfo{person}{Manjunath Kudlur}.} \bibinfo{year}{2017}\natexlab{}.
\newblock \showarticletitle{A Learned Representation For Artistic Style}. In \bibinfo{booktitle}{\emph{International Conference on Learning Representations}}.
\newblock


\bibitem[Frenkel et~al\mbox{.}(2024)]%
        {frenkel2024blora}
\bibfield{author}{\bibinfo{person}{Yarden Frenkel}, \bibinfo{person}{Yael Vinker}, \bibinfo{person}{Ariel Shamir}, {and} \bibinfo{person}{Daniel Cohen-Or}.} \bibinfo{year}{2024}\natexlab{}.
\newblock \showarticletitle{Implicit Style-Content Separation using B-LoRA}.
\newblock \bibinfo{journal}{\emph{arXiv preprint arXiv:2403.14572}} (\bibinfo{year}{2024}).
\newblock


\bibitem[Gal et~al\mbox{.}(2023)]%
        {gal2022textual}
\bibfield{author}{\bibinfo{person}{Rinon Gal}, \bibinfo{person}{Yuval Alaluf}, \bibinfo{person}{Yuval Atzmon}, \bibinfo{person}{Or Patashnik}, \bibinfo{person}{Amit~H. Bermano}, \bibinfo{person}{Gal Chechik}, {and} \bibinfo{person}{Daniel Cohen-Or}.} \bibinfo{year}{2023}\natexlab{}.
\newblock \showarticletitle{An Image is Worth One Word: Personalizing Text-to-Image Generation using Textual Inversion}. In \bibinfo{booktitle}{\emph{International Conference on Learning Representations}}.
\newblock


\bibitem[Gao et~al\mbox{.}(2024)]%
        {gao2024genesistex}
\bibfield{author}{\bibinfo{person}{Chenjian Gao}, \bibinfo{person}{Boyan Jiang}, \bibinfo{person}{Xinghui Li}, \bibinfo{person}{Yingpeng Zhang}, {and} \bibinfo{person}{Qian Yu}.} \bibinfo{year}{2024}\natexlab{}.
\newblock \showarticletitle{GenesisTex: Adapting Image Denoising Diffusion to Texture Space}. In \bibinfo{booktitle}{\emph{Proceedings of the IEEE/CVF Conference on Computer Vision and Pattern Recognition, {CVPR} 2024}}. \bibinfo{pages}{4620--4629}.
\newblock


\bibitem[Gatys et~al\mbox{.}(2015)]%
        {gatys2015texture}
\bibfield{author}{\bibinfo{person}{LA Gatys}, \bibinfo{person}{AS Ecker}, {and} \bibinfo{person}{M Bethge}.} \bibinfo{year}{2015}\natexlab{}.
\newblock \showarticletitle{Texture Synthesis Using Convolutional Neural Networks}. In \bibinfo{booktitle}{\emph{Advances in Neural Information Processing Systems 28: Annual Conference on Neural Information Processing Systems 2015}}. \bibinfo{pages}{262--270}.
\newblock


\bibitem[Gatys et~al\mbox{.}(2016a)]%
        {gatys2015neural}
\bibfield{author}{\bibinfo{person}{Leon Gatys}, \bibinfo{person}{Alexander Ecker}, {and} \bibinfo{person}{Matthias Bethge}.} \bibinfo{year}{2016}\natexlab{a}.
\newblock \showarticletitle{A Neural Algorithm of Artistic Style}.
\newblock \bibinfo{journal}{\emph{Journal of Vision}} \bibinfo{volume}{16}, \bibinfo{number}{12} (\bibinfo{year}{2016}), \bibinfo{pages}{326--326}.
\newblock


\bibitem[Gatys et~al\mbox{.}(2016b)]%
        {gatys2016imagea}
\bibfield{author}{\bibinfo{person}{Leon~A. Gatys}, \bibinfo{person}{Alexander~S. Ecker}, {and} \bibinfo{person}{Matthias Bethge}.} \bibinfo{year}{2016}\natexlab{b}.
\newblock \showarticletitle{Image Style Transfer Using Convolutional Neural Networks}. In \bibinfo{booktitle}{\emph{2016 {IEEE} Conference on Computer Vision and Pattern Recognition, {CVPR} 2016}}. \bibinfo{pages}{2414--2423}.
\newblock


\bibitem[Gu et~al\mbox{.}(2018)]%
        {gu2018arbitrary}
\bibfield{author}{\bibinfo{person}{Shuyang Gu}, \bibinfo{person}{Congliang Chen}, \bibinfo{person}{Jing Liao}, {and} \bibinfo{person}{Lu Yuan}.} \bibinfo{year}{2018}\natexlab{}.
\newblock \showarticletitle{Arbitrary Style Transfer with Deep Feature Reshuffle}. In \bibinfo{booktitle}{\emph{2018 {IEEE/CVF} Conference on Computer Vision and Pattern Recognition, {CVPR} 2018}}. \bibinfo{pages}{8222--8231}.
\newblock


\bibitem[Guo et~al\mbox{.}(2023b)]%
        {guo2024decorate3D}
\bibfield{author}{\bibinfo{person}{Yanhui Guo}, \bibinfo{person}{Xinxin Zuo}, \bibinfo{person}{Peng Dai}, \bibinfo{person}{Juwei Lu}, \bibinfo{person}{Xiaolin Wu}, \bibinfo{person}{Youliang Yan}, \bibinfo{person}{Songcen Xu}, \bibinfo{person}{Xiaofei Wu}, {et~al\mbox{.}}} \bibinfo{year}{2023}\natexlab{b}.
\newblock \showarticletitle{Decorate3D: Text-Driven High-Quality Texture Generation for Mesh Decoration in the Wild}. In \bibinfo{booktitle}{\emph{Advances in Neural Information Processing Systems, {NeurIPS 2023}}}, Vol.~\bibinfo{volume}{36}. \bibinfo{pages}{36664--36676}.
\newblock


\bibitem[Guo et~al\mbox{.}(2023a)]%
        {threestudio2023}
\bibfield{author}{\bibinfo{person}{Yuan-Chen Guo}, \bibinfo{person}{Ying-Tian Liu}, \bibinfo{person}{Ruizhi Shao}, \bibinfo{person}{Christian Laforte}, \bibinfo{person}{Vikram Voleti}, \bibinfo{person}{Guan Luo}, \bibinfo{person}{Chia-Hao Chen}, \bibinfo{person}{Zi-Xin Zou}, \bibinfo{person}{Chen Wang}, \bibinfo{person}{Yan-Pei Cao}, {and} \bibinfo{person}{Song-Hai Zhang}.} \bibinfo{year}{2023}\natexlab{a}.
\newblock \bibinfo{title}{threestudio: A unified framework for 3D content generation}.
\newblock \bibinfo{howpublished}{\url{https://github.com/threestudio-project/threestudio}}.
\newblock


\bibitem[He et~al\mbox{.}(2024)]%
        {he2024freestyle}
\bibfield{author}{\bibinfo{person}{Feihong He}, \bibinfo{person}{Gang Li}, \bibinfo{person}{Mengyuan Zhang}, \bibinfo{person}{Leilei Yan}, \bibinfo{person}{Lingyu Si}, {and} \bibinfo{person}{Fanzhang Li}.} \bibinfo{year}{2024}\natexlab{}.
\newblock \showarticletitle{Freestyle: Free lunch for text-guided style transfer using diffusion models}.
\newblock \bibinfo{journal}{\emph{arXiv preprint arXiv:2401.15636}} (\bibinfo{year}{2024}).
\newblock


\bibitem[Hertz et~al\mbox{.}(2024)]%
        {hertz2024style}
\bibfield{author}{\bibinfo{person}{Amir Hertz}, \bibinfo{person}{Andrey Voynov}, \bibinfo{person}{Shlomi Fruchter}, {and} \bibinfo{person}{Daniel {Cohen-Or}}.} \bibinfo{year}{2024}\natexlab{}.
\newblock \showarticletitle{Style Aligned Image Generation via Shared Attention}.
\newblock \bibinfo{journal}{\emph{arXiv preprint arXiv:2312.02133}} (\bibinfo{year}{2024}).
\newblock


\bibitem[Hessel et~al\mbox{.}(2021)]%
        {hessel2021clipscore}
\bibfield{author}{\bibinfo{person}{Jack Hessel}, \bibinfo{person}{Ari Holtzman}, \bibinfo{person}{Maxwell Forbes}, \bibinfo{person}{Ronan~Le Bras}, {and} \bibinfo{person}{Yejin Choi}.} \bibinfo{year}{2021}\natexlab{}.
\newblock \showarticletitle{CLIPScore: {A} Reference-free Evaluation Metric for Image Captioning}. In \bibinfo{booktitle}{\emph{Proceedings of the 2021 Conference on Empirical Methods in Natural Language Processing, {EMNLP} 2021}}. \bibinfo{pages}{7514--7528}.
\newblock


\bibitem[Ho et~al\mbox{.}(2020)]%
        {ho2020denoising}
\bibfield{author}{\bibinfo{person}{Jonathan Ho}, \bibinfo{person}{Ajay Jain}, {and} \bibinfo{person}{Pieter Abbeel}.} \bibinfo{year}{2020}\natexlab{}.
\newblock \showarticletitle{Denoising diffusion probabilistic models}. In \bibinfo{booktitle}{\emph{Advances in Neural Information Processing Systems, {NeurIPS} 2020}}, Vol.~\bibinfo{volume}{33}. \bibinfo{pages}{6840--6851}.
\newblock


\bibitem[Ho and Salimans(2021)]%
        {ho2022classifier}
\bibfield{author}{\bibinfo{person}{Jonathan Ho} {and} \bibinfo{person}{Tim Salimans}.} \bibinfo{year}{2021}\natexlab{}.
\newblock \showarticletitle{Classifier-Free Diffusion Guidance}. In \bibinfo{booktitle}{\emph{NeurIPS 2021 Workshop on Deep Generative Models and Downstream Applications}}.
\newblock


\bibitem[H{\"{o}}llein et~al\mbox{.}(2022)]%
        {hollein2022stylemesh}
\bibfield{author}{\bibinfo{person}{Lukas H{\"{o}}llein}, \bibinfo{person}{Justin Johnson}, {and} \bibinfo{person}{Matthias Nie{\ss}ner}.} \bibinfo{year}{2022}\natexlab{}.
\newblock \showarticletitle{StyleMesh: Style Transfer for Indoor 3D Scene Reconstructions}. In \bibinfo{booktitle}{\emph{{IEEE/CVF} Conference on Computer Vision and Pattern Recognition, {CVPR} 2022}}. \bibinfo{publisher}{{IEEE}}, \bibinfo{pages}{6188--6198}.
\newblock


\bibitem[Hu et~al\mbox{.}(2022)]%
        {hu2022lora}
\bibfield{author}{\bibinfo{person}{Edward~J Hu}, \bibinfo{person}{Yelong Shen}, \bibinfo{person}{Phillip Wallis}, \bibinfo{person}{Zeyuan Allen-Zhu}, \bibinfo{person}{Yuanzhi Li}, \bibinfo{person}{Shean Wang}, \bibinfo{person}{Lu Wang}, {and} \bibinfo{person}{Weizhu Chen}.} \bibinfo{year}{2022}\natexlab{}.
\newblock \showarticletitle{Lo{RA}: Low-Rank Adaptation of Large Language Models}. In \bibinfo{booktitle}{\emph{International Conference on Learning Representations}}.
\newblock


\bibitem[Huang et~al\mbox{.}(2021)]%
        {huang_2021_3D_scene_stylization}
\bibfield{author}{\bibinfo{person}{Hsin-Ping Huang}, \bibinfo{person}{Hung-Yu Tseng}, \bibinfo{person}{Saurabh Saini}, \bibinfo{person}{Maneesh Singh}, {and} \bibinfo{person}{Ming-Hsuan Yang}.} \bibinfo{year}{2021}\natexlab{}.
\newblock \showarticletitle{Learning to stylize novel views}. In \bibinfo{booktitle}{\emph{2021 {IEEE/CVF} International Conference on Computer Vision, {ICCV} 2021}}. \bibinfo{publisher}{{IEEE}}, \bibinfo{pages}{13849--13858}.
\newblock


\bibitem[Huang and Belongie(2017)]%
        {huang2017arbitrary}
\bibfield{author}{\bibinfo{person}{Xun Huang} {and} \bibinfo{person}{Serge Belongie}.} \bibinfo{year}{2017}\natexlab{}.
\newblock \showarticletitle{Arbitrary Style Transfer in Real-Time with Adaptive Instance Normalization}. In \bibinfo{booktitle}{\emph{2017 {IEEE} International Conference on Computer Vision, {ICCV} 2017}}. \bibinfo{publisher}{{IEEE}}, \bibinfo{pages}{1510--1519}.
\newblock


\bibitem[Huang et~al\mbox{.}(2022)]%
        {huang2022stylizednerf}
\bibfield{author}{\bibinfo{person}{Yi-Hua Huang}, \bibinfo{person}{Yue He}, \bibinfo{person}{Yu-Jie Yuan}, \bibinfo{person}{Yu-Kun Lai}, {and} \bibinfo{person}{Lin Gao}.} \bibinfo{year}{2022}\natexlab{}.
\newblock \showarticletitle{StylizedNeRF: Consistent 3D Scene Stylization as Stylized NeRF via 2D-3D Mutual Learning}. In \bibinfo{booktitle}{\emph{2022 {IEEE/CVF} Conference on Computer Vision and Pattern Recognition, {CVPR} 2022}}. \bibinfo{pages}{18321--18331}.
\newblock


\bibitem[Jeong et~al\mbox{.}(2024)]%
        {jeong2024visual}
\bibfield{author}{\bibinfo{person}{Jaeseok Jeong}, \bibinfo{person}{Junho Kim}, \bibinfo{person}{Yunjey Choi}, \bibinfo{person}{Gayoung Lee}, {and} \bibinfo{person}{Youngjung Uh}.} \bibinfo{year}{2024}\natexlab{}.
\newblock \showarticletitle{Visual Style Prompting with Swapping Self-Attention}.
\newblock \bibinfo{journal}{\emph{arXiv preprint arXiv:2402.12974}} (\bibinfo{year}{2024}).
\newblock


\bibitem[Johnson et~al\mbox{.}(2016)]%
        {johnson2016perceptual}
\bibfield{author}{\bibinfo{person}{Justin Johnson}, \bibinfo{person}{Alexandre Alahi}, {and} \bibinfo{person}{Li Fei{-}Fei}.} \bibinfo{year}{2016}\natexlab{}.
\newblock \showarticletitle{Perceptual Losses for Real-Time Style Transfer and Super-Resolution}. In \bibinfo{booktitle}{\emph{Computer Vision - {ECCV} 2016 - 14th European Conference}} \emph{(\bibinfo{series}{Lecture Notes in Computer Science}, Vol.~\bibinfo{volume}{9906})}. \bibinfo{pages}{694--711}.
\newblock


\bibitem[Kato et~al\mbox{.}(2018)]%
        {kato2018renderer}
\bibfield{author}{\bibinfo{person}{Hiroharu Kato}, \bibinfo{person}{Yoshitaka Ushiku}, {and} \bibinfo{person}{Tatsuya Harada}.} \bibinfo{year}{2018}\natexlab{}.
\newblock \showarticletitle{Neural 3D Mesh Renderer}. In \bibinfo{booktitle}{\emph{2018 {IEEE} Conference on Computer Vision and Pattern Recognition, {CVPR} 2018}}. \bibinfo{publisher}{Computer Vision Foundation / {IEEE} Computer Society}, \bibinfo{pages}{3907--3916}.
\newblock


\bibitem[Kolkin et~al\mbox{.}(2022)]%
        {kolkin2022neural}
\bibfield{author}{\bibinfo{person}{Nicholas Kolkin}, \bibinfo{person}{Michal Kucera}, \bibinfo{person}{Sylvain Paris}, \bibinfo{person}{Daniel Sykora}, \bibinfo{person}{Eli Shechtman}, {and} \bibinfo{person}{Greg Shakhnarovich}.} \bibinfo{year}{2022}\natexlab{}.
\newblock \showarticletitle{Neural Neighbor Style Transfer}.
\newblock \bibinfo{journal}{\emph{arXiv preprint arXiv:2203.13215}} (\bibinfo{year}{2022}).
\newblock


\bibitem[Kotovenko et~al\mbox{.}(2019)]%
        {kotovenko2019iccv}
\bibfield{author}{\bibinfo{person}{adn Sanakoyeu~Artsiom Kotovenko, Dmytro}, \bibinfo{person}{Sabine Lang}, {and} \bibinfo{person}{Bj\"orn Ommer}.} \bibinfo{year}{2019}\natexlab{}.
\newblock \showarticletitle{Content and Style Disentanglement for Artistic Style Transfer}. In \bibinfo{booktitle}{\emph{2019 {IEEE/CVF} International Conference on Computer Vision, {ICCV} 2019}}. \bibinfo{publisher}{{IEEE}}, \bibinfo{pages}{4421--4430}.
\newblock


\bibitem[Le et~al\mbox{.}(2023)]%
        {le2023euclidreamer}
\bibfield{author}{\bibinfo{person}{Cindy Le}, \bibinfo{person}{Congrui Hetang}, \bibinfo{person}{Ang Cao}, {and} \bibinfo{person}{Yihui He}.} \bibinfo{year}{2023}\natexlab{}.
\newblock \showarticletitle{EucliDreamer: Fast and High-Quality Texturing for 3D Models with Stable Diffusion Depth}.
\newblock \bibinfo{journal}{\emph{arXiv preprint arXiv:2311.15573}} (\bibinfo{year}{2023}).
\newblock


\bibitem[Li et~al\mbox{.}(2017)]%
        {WCT-NIPS-2017}
\bibfield{author}{\bibinfo{person}{Yijun Li}, \bibinfo{person}{Chen Fang}, \bibinfo{person}{Jimei Yang}, \bibinfo{person}{Zhaowen Wang}, \bibinfo{person}{Xin Lu}, {and} \bibinfo{person}{Ming-Hsuan Yang}.} \bibinfo{year}{2017}\natexlab{}.
\newblock \showarticletitle{Universal Style Transfer via Feature Transforms}. In \bibinfo{booktitle}{\emph{Proceedings of the 31st International Conference on Neural Information Processing Systems}} \emph{(\bibinfo{series}{NIPS'17})}. \bibinfo{pages}{385–395}.
\newblock


\bibitem[Liang et~al\mbox{.}(2024)]%
        {EnVision2023luciddreamer}
\bibfield{author}{\bibinfo{person}{Yixun Liang}, \bibinfo{person}{Xin Yang}, \bibinfo{person}{Jiantao Lin}, \bibinfo{person}{Haodong Li}, \bibinfo{person}{Xiaogang Xu}, {and} \bibinfo{person}{Yingcong Chen}.} \bibinfo{year}{2024}\natexlab{}.
\newblock \showarticletitle{LucidDreamer: Towards High-Fidelity Text-to-3D Generation via Interval Score Matching}. In \bibinfo{booktitle}{\emph{Proceedings of the IEEE/CVF Conference on Computer Vision and Pattern Recognition, {CVPR} 2024}}. \bibinfo{pages}{6517--6526}.
\newblock


\bibitem[Liu et~al\mbox{.}(2023c)]%
        {liu2023stylerf}
\bibfield{author}{\bibinfo{person}{Kunhao Liu}, \bibinfo{person}{Fangneng Zhan}, \bibinfo{person}{Yiwen Chen}, \bibinfo{person}{Jiahui Zhang}, \bibinfo{person}{Yingchen Yu}, \bibinfo{person}{Abdulmotaleb El~Saddik}, \bibinfo{person}{Shijian Lu}, {and} \bibinfo{person}{Eric~P Xing}.} \bibinfo{year}{2023}\natexlab{c}.
\newblock \showarticletitle{StyleRF: Zero-Shot 3D Style Transfer of Neural Radiance Fields}. In \bibinfo{booktitle}{\emph{2023 {IEEE/CVF} Conference on Computer Vision and Pattern Recognition, {CVPR} 2023}}. \bibinfo{pages}{8338--8348}.
\newblock


\bibitem[Liu et~al\mbox{.}(2023a)]%
        {liu2023syncdreamer}
\bibfield{author}{\bibinfo{person}{Yuan Liu}, \bibinfo{person}{Cheng Lin}, \bibinfo{person}{Zijiao Zeng}, \bibinfo{person}{Xiaoxiao Long}, \bibinfo{person}{Lingjie Liu}, \bibinfo{person}{Taku Komura}, {and} \bibinfo{person}{Wenping Wang}.} \bibinfo{year}{2023}\natexlab{a}.
\newblock \showarticletitle{SyncDreamer: Learning to Generate Multiview-consistent Images from a Single-view Image}.
\newblock \bibinfo{journal}{\emph{arXiv preprint arXiv:2309.03453}} (\bibinfo{year}{2023}).
\newblock


\bibitem[Liu et~al\mbox{.}(2023b)]%
        {liu2023text}
\bibfield{author}{\bibinfo{person}{Yuxin Liu}, \bibinfo{person}{Minshan Xie}, \bibinfo{person}{Hanyuan Liu}, {and} \bibinfo{person}{Tien-Tsin Wong}.} \bibinfo{year}{2023}\natexlab{b}.
\newblock \showarticletitle{Text-Guided Texturing by Synchronized Multi-View Diffusion}.
\newblock \bibinfo{journal}{\emph{arXiv preprint arXiv:2311.12891}} (\bibinfo{year}{2023}).
\newblock


\bibitem[Liu et~al\mbox{.}(2024)]%
        {liu2024texdreamer}
\bibfield{author}{\bibinfo{person}{Yufei Liu}, \bibinfo{person}{Junwei Zhu}, \bibinfo{person}{Junshu Tang}, \bibinfo{person}{Shijie Zhang}, \bibinfo{person}{Jiangning Zhang}, \bibinfo{person}{Weijian Cao}, \bibinfo{person}{Chengjie Wang}, \bibinfo{person}{Yunsheng Wu}, {and} \bibinfo{person}{Dongjin Huang}.} \bibinfo{year}{2024}\natexlab{}.
\newblock \showarticletitle{TexDreamer: Towards Zero-Shot High-Fidelity 3D Human Texture Generation}.
\newblock \bibinfo{journal}{\emph{arXiv preprint arXiv:2403.12906}} (\bibinfo{year}{2024}).
\newblock


\bibitem[Metzer et~al\mbox{.}(2023)]%
        {metzer2023latent}
\bibfield{author}{\bibinfo{person}{Gal Metzer}, \bibinfo{person}{Elad Richardson}, \bibinfo{person}{Or Patashnik}, \bibinfo{person}{Raja Giryes}, {and} \bibinfo{person}{Daniel Cohen-Or}.} \bibinfo{year}{2023}\natexlab{}.
\newblock \showarticletitle{Latent-NeRF for Shape-Guided Generation of 3D Shapes and Textures}. In \bibinfo{booktitle}{\emph{{IEEE/CVF} Conference on Computer Vision and Pattern Recognition, CVPR 2023}}. \bibinfo{pages}{12663--12673}.
\newblock


\bibitem[Mu et~al\mbox{.}(2022)]%
        {mu20213D}
\bibfield{author}{\bibinfo{person}{Fangzhou Mu}, \bibinfo{person}{Jian Wang}, \bibinfo{person}{Yicheng Wu}, {and} \bibinfo{person}{Yin Li}.} \bibinfo{year}{2022}\natexlab{}.
\newblock \showarticletitle{3D Photo Stylization: Learning to Generate Stylized Novel Views from a Single Image}. In \bibinfo{booktitle}{\emph{2022 {IEEE/CVF} Conference on Computer Vision and Pattern Recognition, {CVPR} 2022}}. \bibinfo{pages}{16252--16261}.
\newblock


\bibitem[M{\"u}ller et~al\mbox{.}(2022)]%
        {muller2022instant}
\bibfield{author}{\bibinfo{person}{Thomas M{\"u}ller}, \bibinfo{person}{Alex Evans}, \bibinfo{person}{Christoph Schied}, {and} \bibinfo{person}{Alexander Keller}.} \bibinfo{year}{2022}\natexlab{}.
\newblock \showarticletitle{Instant neural graphics primitives with a multiresolution hash encoding}.
\newblock \bibinfo{journal}{\emph{ACM Transactions on Graphics (ToG)}} \bibinfo{volume}{41}, \bibinfo{number}{4}, Article \bibinfo{articleno}{102} (\bibinfo{year}{2022}), \bibinfo{numpages}{15}~pages.
\newblock


\bibitem[Munkberg et~al\mbox{.}(2022)]%
        {Munkberg_2022_CVPR}
\bibfield{author}{\bibinfo{person}{Jacob Munkberg}, \bibinfo{person}{Jon Hasselgren}, \bibinfo{person}{Tianchang Shen}, \bibinfo{person}{Jun Gao}, \bibinfo{person}{Wenzheng Chen}, \bibinfo{person}{Alex Evans}, \bibinfo{person}{Thomas M\"uller}, {and} \bibinfo{person}{Sanja Fidler}.} \bibinfo{year}{2022}\natexlab{}.
\newblock \showarticletitle{Extracting Triangular 3D Models, Materials, and Lighting From Images}. In \bibinfo{booktitle}{\emph{2022 {IEEE/CVF} Conference on Computer Vision and Pattern Recognition, {CVPR} 2022}}. \bibinfo{pages}{8270--8280}.
\newblock


\bibitem[{Nguyen-Phuoc} et~al\mbox{.}(2022)]%
        {nguyen-phuoc2022snerf}
\bibfield{author}{\bibinfo{person}{Thu {Nguyen-Phuoc}}, \bibinfo{person}{Feng Liu}, {and} \bibinfo{person}{Lei Xiao}.} \bibinfo{year}{2022}\natexlab{}.
\newblock \showarticletitle{SNeRF: Stylized Neural Implicit Representations for 3D Scenes}.
\newblock \bibinfo{journal}{\emph{arXiv preprint arXiv:2207.02363}} (\bibinfo{year}{2022}).
\newblock


\bibitem[Park and Lee(2019)]%
        {park2019arbitrary}
\bibfield{author}{\bibinfo{person}{Dae~Young Park} {and} \bibinfo{person}{Kwang~Hee Lee}.} \bibinfo{year}{2019}\natexlab{}.
\newblock \showarticletitle{Arbitrary Style Transfer With Style-Attentional Networks}. In \bibinfo{booktitle}{\emph{2019 {IEEE/CVF} Conference on Computer Vision and Pattern Recognition, {CVPR} 2019}}. \bibinfo{pages}{5873--5881}.
\newblock


\bibitem[Poole et~al\mbox{.}(2022)]%
        {poole2022dreamfusion}
\bibfield{author}{\bibinfo{person}{Ben Poole}, \bibinfo{person}{Ajay Jain}, \bibinfo{person}{Jonathan~T. Barron}, {and} \bibinfo{person}{Ben Mildenhall}.} \bibinfo{year}{2022}\natexlab{}.
\newblock \showarticletitle{DreamFusion: Text-to-3D using 2D Diffusion}. In \bibinfo{booktitle}{\emph{International Conference on Learning Representations}}.
\newblock


\bibitem[Qi et~al\mbox{.}(2024)]%
        {qi2024deadiff}
\bibfield{author}{\bibinfo{person}{Tianhao Qi}, \bibinfo{person}{Shancheng Fang}, \bibinfo{person}{Yanze Wu}, \bibinfo{person}{Hongtao Xie}, \bibinfo{person}{Jiawei Liu}, \bibinfo{person}{Lang Chen}, \bibinfo{person}{Qian He}, {and} \bibinfo{person}{Yongdong Zhang}.} \bibinfo{year}{2024}\natexlab{}.
\newblock \showarticletitle{DEADiff: An Efficient Stylization Diffusion Model with Disentangled Representations}.
\newblock \bibinfo{journal}{\emph{arXiv preprint arXiv:2403.06951}} (\bibinfo{year}{2024}).
\newblock


\bibitem[Richardson et~al\mbox{.}(2023)]%
        {richardson2023texture}
\bibfield{author}{\bibinfo{person}{Elad Richardson}, \bibinfo{person}{Gal Metzer}, \bibinfo{person}{Yuval Alaluf}, \bibinfo{person}{Raja Giryes}, {and} \bibinfo{person}{Daniel Cohen-Or}.} \bibinfo{year}{2023}\natexlab{}.
\newblock \showarticletitle{Texture: Text-guided texturing of 3d shapes}. In \bibinfo{booktitle}{\emph{ACM SIGGRAPH 2023 Conference Proceedings}} (Los Angeles, CA, USA) \emph{(\bibinfo{series}{SIGGRAPH '23})}. \bibinfo{publisher}{Association for Computing Machinery}, \bibinfo{address}{New York, NY, USA}, Article \bibinfo{articleno}{54}, \bibinfo{numpages}{11}~pages.
\newblock
\urldef\tempurl%
\url{https://doi.org/10.1145/3588432.3591503}
\showDOI{\tempurl}


\bibitem[Rombach et~al\mbox{.}(2022)]%
        {rombach2021highresolution}
\bibfield{author}{\bibinfo{person}{Robin Rombach}, \bibinfo{person}{Andreas Blattmann}, \bibinfo{person}{Dominik Lorenz}, \bibinfo{person}{Patrick Esser}, {and} \bibinfo{person}{Bj{\"{o}}rn Ommer}.} \bibinfo{year}{2022}\natexlab{}.
\newblock \showarticletitle{High-Resolution Image Synthesis with Latent Diffusion Models}. In \bibinfo{booktitle}{\emph{{IEEE/CVF} Conference on Computer Vision and Pattern Recognition, {CVPR} 2022}}. \bibinfo{publisher}{{IEEE}}, \bibinfo{pages}{10674--10685}.
\newblock


\bibitem[Ruiz et~al\mbox{.}(2023)]%
        {ruiz2022dreambooth}
\bibfield{author}{\bibinfo{person}{Nataniel Ruiz}, \bibinfo{person}{Yuanzhen Li}, \bibinfo{person}{Varun Jampani}, \bibinfo{person}{Yael Pritch}, \bibinfo{person}{Michael Rubinstein}, {and} \bibinfo{person}{Kfir Aberman}.} \bibinfo{year}{2023}\natexlab{}.
\newblock \showarticletitle{DreamBooth: Fine Tuning Text-to-Image Diffusion Models for Subject-Driven Generation}. In \bibinfo{booktitle}{\emph{2023 {IEEE/CVF} Conference on Computer Vision and Pattern Recognition, {CVPR} 2023}}. \bibinfo{pages}{22500--22510}.
\newblock


\bibitem[Shah et~al\mbox{.}(2023)]%
        {shah2023ZipLoRA}
\bibfield{author}{\bibinfo{person}{Viraj Shah}, \bibinfo{person}{Nataniel Ruiz}, \bibinfo{person}{Forrester Cole}, \bibinfo{person}{Erika Lu}, \bibinfo{person}{Svetlana Lazebnik}, \bibinfo{person}{Yuanzhen Li}, {and} \bibinfo{person}{Varun Jampani}.} \bibinfo{year}{2023}\natexlab{}.
\newblock \showarticletitle{ZipLoRA: Any Subject in Any Style by Effectively Merging LoRAs}.
\newblock \bibinfo{journal}{\emph{arXiv preprint arxiv:2311.13600}} (\bibinfo{year}{2023}).
\newblock


\bibitem[Siddiqui et~al\mbox{.}(2022)]%
        {siddiqui2022texturify}
\bibfield{author}{\bibinfo{person}{Yawar Siddiqui}, \bibinfo{person}{Justus Thies}, \bibinfo{person}{Fangchang Ma}, \bibinfo{person}{Qi Shan}, \bibinfo{person}{Matthias Nie{\ss}ner}, {and} \bibinfo{person}{Angela Dai}.} \bibinfo{year}{2022}\natexlab{}.
\newblock \showarticletitle{Texturify: Generating Textures on 3D Shape Surfaces}. In \bibinfo{booktitle}{\emph{Computer Vision – ECCV 2022: 17th European Conference}} (Tel Aviv, Israel). \bibinfo{pages}{72–88}.
\newblock


\bibitem[{Sketchfab}({[n.\,d.]})]%
        {Sketchfab}
\bibfield{author}{\bibinfo{person}{{Sketchfab}}.} \bibinfo{year}{[n.\,d.]}\natexlab{}.
\newblock \bibinfo{title}{Sketchfab - The best 3D viewer on the web}.
\newblock
\newblock
\urldef\tempurl%
\url{https://www.sketchfab.com}
\showURL{%
\tempurl}


\bibitem[Sohn et~al\mbox{.}(2024)]%
        {sohn2023styledrop}
\bibfield{author}{\bibinfo{person}{Kihyuk Sohn}, \bibinfo{person}{Nataniel Ruiz}, \bibinfo{person}{Kimin Lee}, \bibinfo{person}{Daniel~Castro Chin}, \bibinfo{person}{Irina Blok}, \bibinfo{person}{Huiwen Chang}, \bibinfo{person}{Jarred Barber}, \bibinfo{person}{Lu Jiang}, \bibinfo{person}{Glenn Entis}, \bibinfo{person}{Yuanzhen Li}, {et~al\mbox{.}}} \bibinfo{year}{2024}\natexlab{}.
\newblock \showarticletitle{StyleDrop: Text-to-Image Generation in Any Style}. In \bibinfo{booktitle}{\emph{Proceedings of the 37th International Conference on Neural Information Processing Systems}}. Article \bibinfo{articleno}{2920}, \bibinfo{numpages}{30}~pages.
\newblock


\bibitem[Song et~al\mbox{.}(2020)]%
        {song2020denoising}
\bibfield{author}{\bibinfo{person}{Jiaming Song}, \bibinfo{person}{Chenlin Meng}, {and} \bibinfo{person}{Stefano Ermon}.} \bibinfo{year}{2020}\natexlab{}.
\newblock \showarticletitle{Denoising Diffusion Implicit Models}. In \bibinfo{booktitle}{\emph{International Conference on Learning Representations}}.
\newblock


\bibitem[Ulyanov et~al\mbox{.}(2016)]%
        {Ulyanov2016TextureNF}
\bibfield{author}{\bibinfo{person}{Dmitry Ulyanov}, \bibinfo{person}{Vadim Lebedev}, \bibinfo{person}{Andrea Vedaldi}, {and} \bibinfo{person}{Victor~S. Lempitsky}.} \bibinfo{year}{2016}\natexlab{}.
\newblock \showarticletitle{Texture Networks: Feed-forward Synthesis of Textures and Stylized Images}.
\newblock \bibinfo{journal}{\emph{arXiv preprint arXiv:1603.03417}} (\bibinfo{year}{2016}).
\newblock


\bibitem[Voynov et~al\mbox{.}(2023)]%
        {voynov2023p+}
\bibfield{author}{\bibinfo{person}{Andrey Voynov}, \bibinfo{person}{Qinghao Chu}, \bibinfo{person}{Daniel Cohen-Or}, {and} \bibinfo{person}{Kfir Aberman}.} \bibinfo{year}{2023}\natexlab{}.
\newblock \showarticletitle{$ P+ $: Extended Textual Conditioning in Text-to-Image Generation}.
\newblock \bibinfo{journal}{\emph{arXiv preprint arXiv:2303.09522}} (\bibinfo{year}{2023}).
\newblock


\bibitem[Wang et~al\mbox{.}(2024)]%
        {wang2024instantstyle}
\bibfield{author}{\bibinfo{person}{Haofan Wang}, \bibinfo{person}{Qixun Wang}, \bibinfo{person}{Xu Bai}, \bibinfo{person}{Zekui Qin}, {and} \bibinfo{person}{Anthony Chen}.} \bibinfo{year}{2024}\natexlab{}.
\newblock \showarticletitle{InstantStyle: Free Lunch towards Style-Preserving in Text-to-Image Generation}.
\newblock \bibinfo{journal}{\emph{arXiv preprint arXiv:2404.02733}} (\bibinfo{year}{2024}).
\newblock


\bibitem[Wang et~al\mbox{.}(2023)]%
        {ye2023styleadapter}
\bibfield{author}{\bibinfo{person}{Zhouxia Wang}, \bibinfo{person}{Xintao Wang}, \bibinfo{person}{Liangbin Xie}, \bibinfo{person}{Zhongang Qi}, \bibinfo{person}{Ying Shan}, \bibinfo{person}{Wenping Wang}, {and} \bibinfo{person}{Ping Luo}.} \bibinfo{year}{2023}\natexlab{}.
\newblock \showarticletitle{StyleAdapter: A Single-Pass LoRA-Free Model for Stylized Image Generation}.
\newblock \bibinfo{journal}{\emph{arXiv preprint arxiv:2309.01770}} (\bibinfo{year}{2023}).
\newblock


\bibitem[Wu et~al\mbox{.}(2024)]%
        {wu2024texro}
\bibfield{author}{\bibinfo{person}{Jinbo Wu}, \bibinfo{person}{Xing Liu}, \bibinfo{person}{Chenming Wu}, \bibinfo{person}{Xiaobo Gao}, \bibinfo{person}{Jialun Liu}, \bibinfo{person}{Xinqi Liu}, \bibinfo{person}{Chen Zhao}, \bibinfo{person}{Haocheng Feng}, \bibinfo{person}{Errui Ding}, {and} \bibinfo{person}{Jingdong Wang}.} \bibinfo{year}{2024}\natexlab{}.
\newblock \showarticletitle{TexRO: Generating Delicate Textures of 3D Models by Recursive Optimization}.
\newblock \bibinfo{journal}{\emph{arXiv preprint arXiv:2403.15009}} (\bibinfo{year}{2024}).
\newblock


\bibitem[Ye et~al\mbox{.}(2023)]%
        {ye2023ip-adapter}
\bibfield{author}{\bibinfo{person}{Hu Ye}, \bibinfo{person}{Jun Zhang}, \bibinfo{person}{Sibo Liu}, \bibinfo{person}{Xiao Han}, {and} \bibinfo{person}{Wei Yang}.} \bibinfo{year}{2023}\natexlab{}.
\newblock \showarticletitle{IP-Adapter: Text Compatible Image Prompt Adapter for Text-to-Image Diffusion Models}.
\newblock \bibinfo{journal}{\emph{arXiv preprint arxiv:2308.06721}} (\bibinfo{year}{2023}).
\newblock


\bibitem[Yeh et~al\mbox{.}(2024)]%
        {yeh2024texturedreamer}
\bibfield{author}{\bibinfo{person}{Yu-Ying Yeh}, \bibinfo{person}{Jia-Bin Huang}, \bibinfo{person}{Changil Kim}, \bibinfo{person}{Lei Xiao}, \bibinfo{person}{Thu Nguyen-Phuoc}, \bibinfo{person}{Numair Khan}, \bibinfo{person}{Cheng Zhang}, \bibinfo{person}{Manmohan Chandraker}, \bibinfo{person}{Carl~S Marshall}, \bibinfo{person}{Zhao Dong}, {et~al\mbox{.}}} \bibinfo{year}{2024}\natexlab{}.
\newblock \showarticletitle{TextureDreamer: Image-guided Texture Synthesis through Geometry-aware Diffusion}.
\newblock \bibinfo{journal}{\emph{arXiv preprint arXiv:2401.09416}} (\bibinfo{year}{2024}).
\newblock


\bibitem[Yin et~al\mbox{.}(2021)]%
        {yin2021_3DStyleNet}
\bibfield{author}{\bibinfo{person}{Kangxue Yin}, \bibinfo{person}{Jun Gao}, \bibinfo{person}{Maria Shugrina}, \bibinfo{person}{Sameh Khamis}, {and} \bibinfo{person}{Sanja Fidler}.} \bibinfo{year}{2021}\natexlab{}.
\newblock \showarticletitle{3DStyleNet: Creating 3D Shapes with Geometric and Texture Style Variations}. In \bibinfo{booktitle}{\emph{2021 {IEEE/CVF} International Conference on Computer Vision, {ICCV} 2021}}. \bibinfo{publisher}{{IEEE}}, \bibinfo{pages}{12436--12445}.
\newblock


\bibitem[Young(2021)]%
        {young2021jpcy}
\bibfield{author}{\bibinfo{person}{Jonathan Young}.} \bibinfo{year}{2021}\natexlab{}.
\newblock \bibinfo{title}{Jpcy/Xatlas}.
\newblock
\newblock
\urldef\tempurl%
\url{https://github.com/jpcy/xatlas.git}
\showURL{%
\tempurl}


\bibitem[Youwang et~al\mbox{.}(2023)]%
        {youwang2023paint}
\bibfield{author}{\bibinfo{person}{Kim Youwang}, \bibinfo{person}{Tae-Hyun Oh}, {and} \bibinfo{person}{Gerard Pons-Moll}.} \bibinfo{year}{2023}\natexlab{}.
\newblock \showarticletitle{Paint-it: Text-to-Texture Synthesis via Deep Convolutional Texture Map Optimization and Physically-Based Rendering}.
\newblock \bibinfo{journal}{\emph{arXiv preprint arXiv:2312.11360}} (\bibinfo{year}{2023}).
\newblock


\bibitem[Zeng et~al\mbox{.}(2023b)]%
        {zeng2023ipdreamer}
\bibfield{author}{\bibinfo{person}{Bohan Zeng}, \bibinfo{person}{Shanglin Li}, \bibinfo{person}{Yutang Feng}, \bibinfo{person}{Hong Li}, \bibinfo{person}{Sicheng Gao}, \bibinfo{person}{Jiaming Liu}, \bibinfo{person}{Huaxia Li}, \bibinfo{person}{Xu Tang}, \bibinfo{person}{Jianzhuang Liu}, {and} \bibinfo{person}{Baochang Zhang}.} \bibinfo{year}{2023}\natexlab{b}.
\newblock \showarticletitle{Ipdreamer: Appearance-controllable 3d object generation with image prompts}.
\newblock \bibinfo{journal}{\emph{arXiv preprint arXiv:2310.05375}} (\bibinfo{year}{2023}).
\newblock


\bibitem[Zeng et~al\mbox{.}(2023a)]%
        {zeng2023paint3D}
\bibfield{author}{\bibinfo{person}{Xianfang Zeng}, \bibinfo{person}{Xin Chen}, \bibinfo{person}{Zhongqi Qi}, \bibinfo{person}{Wen Liu}, \bibinfo{person}{Zibo Zhao}, \bibinfo{person}{Zhibin Wang}, \bibinfo{person}{Bin Fu}, \bibinfo{person}{Yong Liu}, {and} \bibinfo{person}{Gang Yu}.} \bibinfo{year}{2023}\natexlab{a}.
\newblock \showarticletitle{Paint3D: Paint Anything 3D with Lighting-Less Texture Diffusion Models}.
\newblock \bibinfo{journal}{\emph{arXiv preprint arXiv:2312.13913}} (\bibinfo{year}{2023}).
\newblock


\bibitem[Zhang et~al\mbox{.}(2024a)]%
        {Zhang2024StylizedGS}
\bibfield{author}{\bibinfo{person}{Dingxi Zhang}, \bibinfo{person}{Zhuoxun Chen}, \bibinfo{person}{Yujian Yuan}, \bibinfo{person}{Fang-Lue Zhang}, \bibinfo{person}{Zhenliang He}, \bibinfo{person}{Shiguang Shan}, {and} \bibinfo{person}{Lin Gao}.} \bibinfo{year}{2024}\natexlab{a}.
\newblock \showarticletitle{StylizedGS: Controllable Stylization for 3D Gaussian Splatting}.
\newblock \bibinfo{journal}{\emph{arXiv preprint arXiv:2404.05220}} (\bibinfo{year}{2024}).
\newblock


\bibitem[Zhang and Dana(2019)]%
        {zhng2019multistyle}
\bibfield{author}{\bibinfo{person}{Hang Zhang} {and} \bibinfo{person}{Kristin Dana}.} \bibinfo{year}{2019}\natexlab{}.
\newblock \showarticletitle{Multi-Style Generative Network for Real-Time Transfer}. In \bibinfo{booktitle}{\emph{Computer Vision – ECCV 2018 Workshops: Munich}}. \bibinfo{pages}{349–365}.
\newblock


\bibitem[Zhang et~al\mbox{.}(2022)]%
        {zhang2022arf}
\bibfield{author}{\bibinfo{person}{Kai Zhang}, \bibinfo{person}{Nick Kolkin}, \bibinfo{person}{Sai Bi}, \bibinfo{person}{Fujun Luan}, \bibinfo{person}{Zexiang Xu}, \bibinfo{person}{Eli Shechtman}, {and} \bibinfo{person}{Noah Snavely}.} \bibinfo{year}{2022}\natexlab{}.
\newblock \showarticletitle{ARF: Artistic Radiance Fields}. In \bibinfo{booktitle}{\emph{Computer Vision – ECCV 2022: 17th European Conference}}. \bibinfo{pages}{717–733}.
\newblock


\bibitem[Zhang et~al\mbox{.}(2023b)]%
        {zhang2023adding}
\bibfield{author}{\bibinfo{person}{Lvmin Zhang}, \bibinfo{person}{Anyi Rao}, {and} \bibinfo{person}{Maneesh Agrawala}.} \bibinfo{year}{2023}\natexlab{b}.
\newblock \showarticletitle{Adding Conditional Control to Text-to-Image Diffusion Models}. In \bibinfo{booktitle}{\emph{{IEEE/CVF} International Conference on Computer Vision, {ICCV} 2023}}. \bibinfo{publisher}{{IEEE}}, \bibinfo{pages}{3813--3824}.
\newblock


\bibitem[Zhang et~al\mbox{.}(2024c)]%
        {zhang2024clay}
\bibfield{author}{\bibinfo{person}{Longwen Zhang}, \bibinfo{person}{Ziyu Wang}, \bibinfo{person}{Qixuan Zhang}, \bibinfo{person}{Qiwei Qiu}, \bibinfo{person}{Anqi Pang}, \bibinfo{person}{Haoran Jiang}, \bibinfo{person}{Wei Yang}, \bibinfo{person}{Lan Xu}, {and} \bibinfo{person}{Jingyi Yu}.} \bibinfo{year}{2024}\natexlab{c}.
\newblock \showarticletitle{CLAY: A Controllable Large-scale Generative Model for Creating High-quality 3D Assets}.
\newblock \bibinfo{journal}{\emph{ACM Trans. Graph.}} \bibinfo{volume}{43}, \bibinfo{number}{4}, Article \bibinfo{articleno}{120} (\bibinfo{year}{2024}), \bibinfo{numpages}{20}~pages.
\newblock
\showISSN{0730-0301}
\urldef\tempurl%
\url{https://doi.org/10.1145/3658146}
\showDOI{\tempurl}


\bibitem[Zhang et~al\mbox{.}(2023a)]%
        {Zhang2023inst}
\bibfield{author}{\bibinfo{person}{Yuxin Zhang}, \bibinfo{person}{Nisha Huang}, \bibinfo{person}{Fan Tang}, \bibinfo{person}{Haibin Huang}, \bibinfo{person}{Chongyang Ma}, \bibinfo{person}{Weiming Dong}, {and} \bibinfo{person}{Changsheng Xu}.} \bibinfo{year}{2023}\natexlab{a}.
\newblock \showarticletitle{Inversion-based Style Transfer with Diffusion Models}. In \bibinfo{booktitle}{\emph{2023 {IEEE/CVF} Conference on Computer Vision and Pattern Recognition, {CVPR} 2023}}. \bibinfo{publisher}{{IEEE}}, \bibinfo{pages}{10146--10156}.
\newblock


\bibitem[Zhang et~al\mbox{.}(2024b)]%
        {zhang2024dreammat}
\bibfield{author}{\bibinfo{person}{Yuqing Zhang}, \bibinfo{person}{Yuan Liu}, \bibinfo{person}{Zhiyu Xie}, \bibinfo{person}{Lei Yang}, \bibinfo{person}{Zhongyuan Liu}, \bibinfo{person}{Mengzhou Yang}, \bibinfo{person}{Runze Zhang}, \bibinfo{person}{Qilong Kou}, \bibinfo{person}{Cheng Lin}, \bibinfo{person}{Wenping Wang}, {and} \bibinfo{person}{Xiaogang Jin}.} \bibinfo{year}{2024}\natexlab{b}.
\newblock \showarticletitle{DreamMat: High-quality PBR Material Generation with Geometry- and Light-aware Diffusion Models}.
\newblock \bibinfo{journal}{\emph{ACM Trans. Graph.}} \bibinfo{volume}{43}, \bibinfo{number}{4}, Article \bibinfo{articleno}{39} (\bibinfo{year}{2024}), \bibinfo{numpages}{18}~pages.
\newblock
\urldef\tempurl%
\url{https://doi.org/10.1145/3658170}
\showDOI{\tempurl}


\end{thebibliography}

\appendix

\section{Appendix}

\revise{
\subsection{Detailed Difference with InstantStyle}
\begin{figure}[h]
  \includegraphics[width=0.99\linewidth]{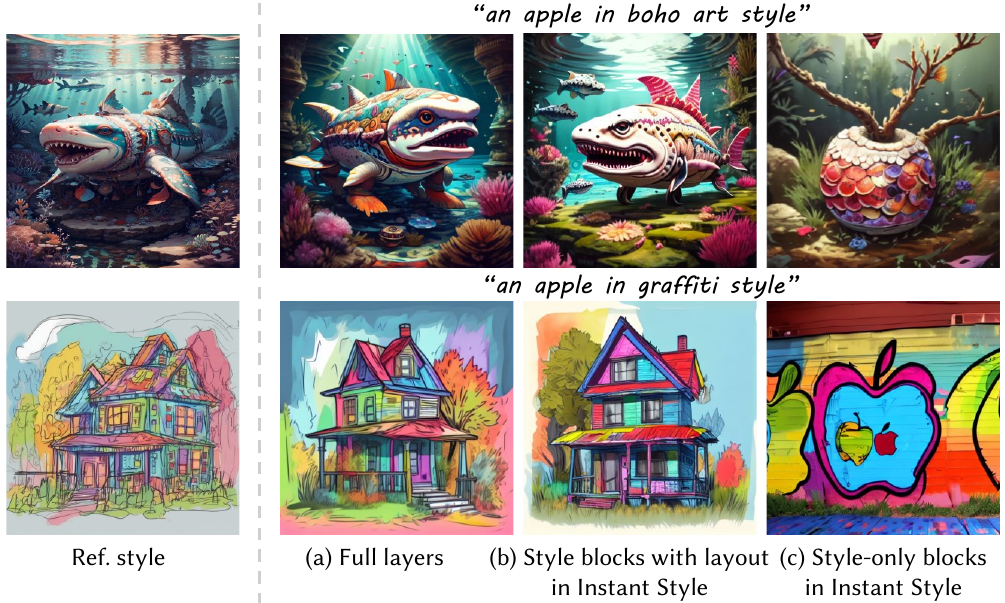}
  \caption{\revise{Results using different types of layers in InstantStyle.} }
  \label{fig:layers_2d}
  \vspace{-3mm}
\end{figure}
InstantStyle~\cite{wang2024instantstyle}  categorizes attention layers that influence style into two types: style-only and spatial layout. In 2D image generation, using full layers can introduce both the reference image's content and style information (see Fig.~\ref{fig:layers_2d} (a)). Employing both the style-only and layout layers may introduce stylistic information as well as spatial structural information (see Fig.~\ref{fig:layers_2d} (b)), whereas only using the style-only layer may result in minor tonal discrepancies (see Fig.~\ref{fig:layers_2d} (c)). In 3D contexts, excessive structural information from layout layers may result in content leakage, and the absence of tonal information from style-only layers can cause severe tonal shifts. Furthermore, InstantStyle uses a simple feature subtraction technique to separate style and content. The style feature is obtained by subtracting the text embedding from the image embedding, resulting in partial content information leakage. 

Unlike their approach, we use InstantStyle's style-only and layout layers, as well as additional layers~\cite{voynov2023p+,agarwal2023image}, to preserve complete style information and avoid tonal shifts. To remove as much structural and content information from the reference image as possible, we use ODCR to extract style features. Furthermore, the content description of the reference image serves as a negative prompt during the distillation process.

%

}

\subsection{Additional Transformer Layers}
\label{sec: Transformer Layers}

The cross-attention layers Instant Style uses for style injection including:
\begin{itemize}
\item down\_blocks.2
\item mid\_block.attention.0
\item up\_block.1
\end{itemize}
In StyleTex, we expand the number of cross-attention layers used for style injection, including:
\begin{itemize}
\item down\_blocks.1.attentions.0
\item All layers in up\_block
\end{itemize}

\begin{figure}[th]
  \includegraphics[width=0.99\linewidth]{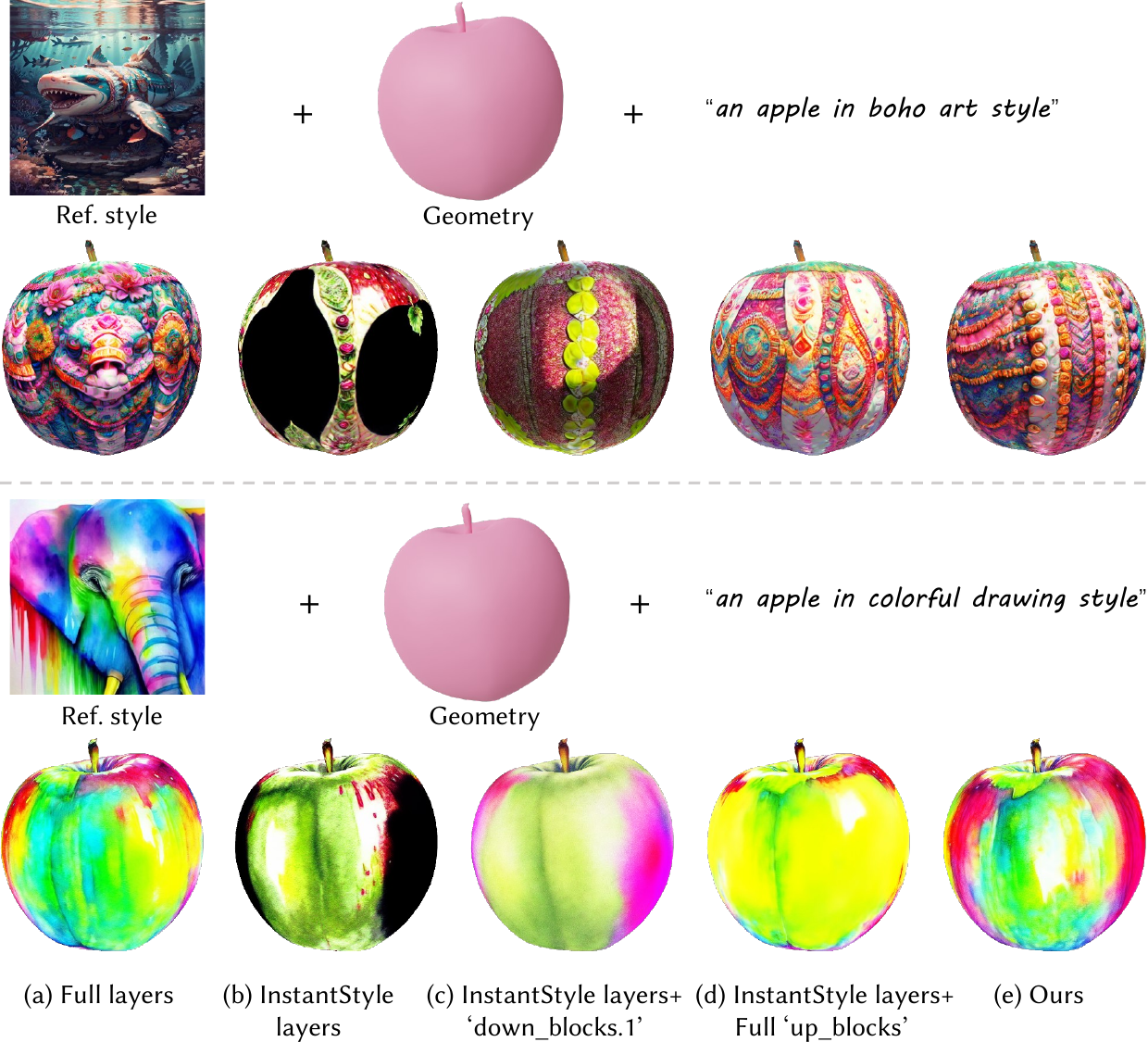}
  \caption{The impact of the additional transformer layers leveraged in our method. }
  \label{fig:layers_3d}
  \vspace{-3mm}
\end{figure}

To evaluate the impact of the additional transformer layers used, we conducted an experiment in which we modified the transformer layers in our full model. The results are presented in Fig. \ref{fig:layers_3d}.
Fig. \ref{fig:layers_3d} (a) demonstrates that injecting style information into all layers results in content leakage issues. Fig. \ref{fig:layers_3d} (b) shows that using only the original injection layer of Instant Style leads to style drift and black areas due to the removal of too many layers in the style injection. By solely adding ``down\_blocks.1.attentions.0'' or ``up\_blocks'', as depicted in Fig. \ref{fig:layers_3d} (c) and (d), respectively, the black area is effectively removed; however, a slight color shift still occurs.
In contrast, using the additional layers as we did in our proposed approach produces results that more closely align with the reference image while avoiding content leakage.

\subsection{Effect of guidance scale}

\begin{figure*}[t]
  \includegraphics[width=0.9\linewidth]{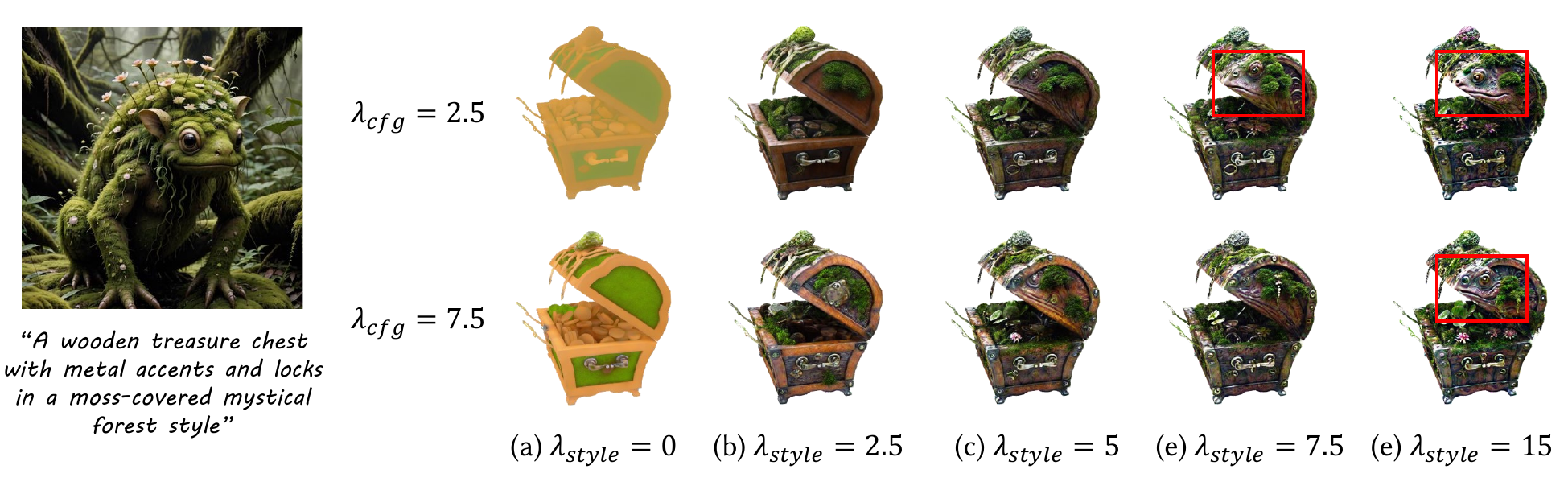}
  \caption{Stylized texture generation with different $\lambda_{cfg}$ and $\lambda_{style}$.}
  \label{fig:scale}
  \vspace{-4mm}
\end{figure*}

In this section, we conduct an investigation into the impact of two hyperparameters: the CFG scale $\lambda_{cfg}$ and the style guidance scale $\lambda_{style}$. As illustrated in Fig.~\ref{fig:scale}, we visualize the influence of both $\lambda_{cfg}$ and $\lambda_{style}$. Our observations indicate that an increase in $\lambda_{style}$ can effectively enhance detail and style guidance. However, if $\lambda_{style}$ becomes excessively high and $\lambda_{cfg}$ is unable to match it, the content text prompt, which serves as a negative prompt in the CFG term, may fail to perform its role adequately, leading to content leakage issues.
It is worth noting that our optimal values for $\lambda_{cfg}$ and $\lambda_{style}$ are suitable for all objects and require no further modification during inference.

\subsection{Implementation Details}
\label{sec:Implementation Details}

\subsubsection{Training Details} 
\label{sec:Training Details}
Our texture generation pipeline is developed in Threestudio~\cite{threestudio2023} with Stable Diffusion 1.5~\cite{rombach2021highresolution}. Our evaluation dataset includes 100 3D models from Objaverse~\cite{deitke2023objaverse} and Sketchfab~\cite{Sketchfab} (see details in Sec. \ref{sec:attribution}). 
The stylistic images for our experiments are derived from the internet or generated by diffusion models (see details in Sec. \ref{sec:attribution}). 
The content text prompts $y_{ref}$ for these style images are obtained via GPT-4~\cite{achiam2023gpt}. 

We optimize the texture field for 2500 iterations using an Adam optimizer with a leaning rate of 0.005. 
During the optimization phase, we employ the pre-trained depth and normal ControlNet~\cite{zhang2023adding} to ensure the alignment of the texture details with the geometry of the input mesh. Both the depth map and normal map are rendered in camera space and subsequently normalized to adhere to ScanNet’s standards~\cite{dai2017scannet}. 
In the main paper, the hyperparameters $\lambda_{cfg}$ in Eq. 7 and $\lambda_{style}$ in Eq. 8 are both set as 7.5.

\subsubsection{Texture Map Extraction} 
\label{sec: Texture Map Extraction}
After obtaining the optimized texture field, we employ a post-processing procedure to ensure the storability, editability, and applicability of the textures across various rendering platforms by transforming the texture field into a texture map with a resolution of $1024^2$. Specifically, similar to ~\cite{Munkberg_2022_CVPR, Chen_2023_ICCV}, we sample the texture field using either the model's inherent UV map or one automatically generated by xatlas~\cite{young2021jpcy}. Furthermore, we apply the UV edge padding technique to fill in the empty regions between UV islands, effectively eliminating unwanted seams.

\subsubsection{Baseline Implementation Details}
\label{sec: Baseline Implementation Details}
In our implementation of TEXTure~\cite{richardson2023texture}, we adhere to its texture-from-image methodology. As TextureDreamer's~\cite{yeh2024texturedreamer} source code is not publicly available, we reproduce their method using threestudio~\cite{threestudio2023}. Due to the absence of specific training details for DreamBooth~\cite{ruiz2022dreambooth} in their publication, we utilize the code and default parameters from the Diffusers library to train DreamBooth with LoRA using a single reference image. IPDreamer~\cite{zeng2023ipdreamer} is a two-stage 3D generation method, with the first stage optimizing geometry and the second stage optimizing appearance. We skip the first stage and feed the input mesh directly to the second stage to optimize the surface color. SyncDreamer~\cite{liu2023syncdreamer} is a method that synthesizes multi-view consistent images based on a given mesh, making it compatible with any 2D image-guided method during the denoising process. Consequently, we employ Instant Style~\cite{wang2024instantstyle} to infuse the style of the reference image.
\subsubsection{Quantitative Evaluation Matrix}
\label{sec: Quantitative Evaluation Matrix}
The quantitative metrics used in our paper are derived from two aspects: alignment with the style of the reference image, and alignment with the text prompts.

\textbf{Gram Matrix Distance.} 
Drawing from traditional 2D style transfer methods~\cite{gatys2015texture,gatys2015neural,johnson2016perceptual}, the squared Frobenius norm of the difference between the Gram matrices of the reference image and the rendered views of the generated textures can be employed to quantify the stylistic divergence:
\begin{equation}
    D_{GM}^{j}=||G^{\phi}_j(I_{ref})-G^{\phi}_j(I_{render})||^2_F ,
\end{equation}
\begin{equation}
    G^{\phi}_j(I)_{c,c'}=\frac{1}{C_j H_j W_j}\sum_{h=1}^{H_j}\sum_{w=1}^{W_j}\phi_j(I)_{h,w,c}\phi_j(I)_{h,w,c'},
\end{equation}
where $\phi_j(I)$ is the activations at the $j$th layer of the VGG network $\phi$ for the input image $I$, which is a feature map of shape $C_j\times H_j\times W_j$.

\textbf{CLIP Score.} CLIP Score~\cite{hessel2021clipscore} is a metric that quantifies the semantic similarity between images and texts. For a rendered view with
visual CLIP embedding $\mathbf{v}$ and a given text prompt with textual CLIP embedding $\mathbf{c}$, we set $w = 2.5$ and compute CLIP Score as:
\begin{equation}
CLIP_s(\mathbf{c},\mathbf{v})=w*max(cos(\mathbf{c},\mathbf{v}),0).
\end{equation}

\subsection{3D Model / Style Image Attribution}
\label{sec:attribution}
In this paper, we use 3D models sourced from the Objaverse~\cite{deitke2023objaverse} and Sketchfab~\cite{Sketchfab} under the Creative Commons Attribution 4.0 International (CC BY 4.0) license. The models are utilized without their original textures to focus solely on the impact of our stylized texture generation method.

Each model used from Sketchfab is attributed as follows:
\begin{itemize}
\item ``\href{https://sketchfab.com/3d-models/baby-animals-statuettes-cadc2617612d47468e92360960583dc9}{Baby Animals Statuettes}'' by Andrei Alexandrescu.
\item``\href{https://sketchfab.com/3d-models/dragon-fruit-52fbf57abe014774bd75a75757ebf6ab}{Dragon Fruit}'' by Andrei Alexandrescu.
\item ``\href{https://sketchfab.com/3d-models/durian-the-king-of-fruits-62cc563e52514fa9b2e3dfdfc09e5377}{Durian The King of Fruits}'' by Laithai.
\item ``\href{https://sketchfab.com/3d-models/backpack-1bb0c9be995747e9abe0831d94066fc8}{Backpack}'' by mickeymoose1204.
\item ``\href{https://sketchfab.com/3d-models/treasure-chest-773a2f35025b4e2e9ac48fd84c16b3ab}{Treasure chest}'' by DailyArt.
\item ``\href{https://sketchfab.com/3d-models/molino-de-viento---windmill-2ea0a5296d4b49dbad71ce1975c0e3ff}{Molino De Viento \_ Windmill}'' by BC-X.
\item ``\href{https://sketchfab.com/3d-models/medivalhousehouse-for-livingmedivalvilage-ba53607959b0476fb719043c406bc245}{MedivalHouse\ |\ house for living\ |\ MedivalVilage}'' by JFred-chill.
\item ``\href{https://sketchfab.com/3d-models/bouddha-statue-photoscanned-2d71e5b04f184ef89130eb26bc726add}{Bouddha Statue Photoscanned}'' by amcgi.
\item ``\href{https://sketchfab.com/3d-models/bunny-c362411a4a744b6bb18ce4ffcf4e7f43}{Bunny}'' by vivienne0716.
\item ``\href{https://sketchfab.com/3d-models/bird-05e5ad206547428699ea8d6c76dfc9a9}{bird}'' by rudolfs.
\item ``\href{https://sketchfab.com/3d-models/vase-rawscan-98a29620a45e47ccb80a75d5416c8255}{Vase .::RAWscan::.}'' by Andrea Spognetta (Spogna).
\item ``\href{https://sketchfab.com/3d-models/leather-wooden-chest-3d-scan-quixel-megascans-36731e079ff044febf94c0fe9ec1d063}{Leather Wooden Chest - 3D scan Quixel Megascans}'' by Guay0.
\item ``\href{https://sketchfab.com/3d-models/stone-2312aa27f10748b699c97ce7e838df19}{Stone}'' by mesropash97.
\item ``\href{https://sketchfab.com/3d-models/stone-2e966dce9db34fecad1290452bdab165}{Stone}'' by Xephira.
\item ``\href{https://sketchfab.com/3d-models/stone-entrance-ea4a511423bd4a1dab2fe32665edd67e}{Stone Entrance}'' by DJMaesen.
\item ``\href{https://sketchfab.com/3d-models/chinese-bridge-ornament-b254f3156a4d45fb841faf6c277719ce}{Chinese Bridge Ornament}'' by artfletch.
\item ``\href{https://sketchfab.com/3d-models/chinese-hall-42672df3ccfd4215b5e4c7f4968c0969}{Chinese Hall}'' by LSDWaterPipe.
\item ``\href{https://sketchfab.com/3d-models/teascroll-clubhouse-scene-e344e3cddaf14f1abacebe09e7dac089}{TeaScroll Clubhouse Scene}'' by Ana\"is Faure.
\item ``\href{https://sketchfab.com/3d-models/hw-xyz-school-hinese-house-d7bdb7a45be248bbb02c515a77384d1a}{Chinese House}'' by GloomyGN.
\item ``\href{https://sketchfab.com/3d-models/chinese-dragon-fan-2efb3988ffcb4bc2bd2a5911d9e764e7}{Chinese Dragon Fan}'' by Mrs. Chief.
\item ``\href{https://sketchfab.com/3d-models/chinese-lacquer-shanxi-console-table-fb09d8719c9b4f81ba5ca2244fcc7588}{Chinese Lacquer Shanxi Console Table}'' by Arts and Materials Lab.
\item ``\href{https://sketchfab.com/3d-models/chinese-cup-97cf49884b3c47b7b930b5bd097ead7b}{chinese cup}'' by Konstantin Morozov.
\item ``\href{https://sketchfab.com/3d-models/teaset-c072fcbce5db4f4e81a499e1cd187432}{Teaset}'' by nuts.
\item ``\href{https://sketchfab.com/3d-models/victorian-cabinet-81291e282b854934a6f1b9f53a900f5f}{victorian Cabinet}'' by lagesnpiet.
\item ``\href{https://sketchfab.com/3d-models/sakura-cherry-blossom-f5e6f5a985ea4fc2a14ee0b4b37572b5}{Sakura Cherry Blossom
}'' by ffish.asia / floraZia.com.
\item ``\href{https://sketchfab.com/3d-models/madrona-invasives-6e1e4cf5c6d74da286cbf425a939c9ce}{Madrona Invasives}'' by dipietron.
\item ``\href{https://sketchfab.com/3d-models/chinese-storehouse-caa4a09b78c344f5a568ce667eccef4c}{Chinese storehouse}'' by LSDWaterPipe.
\item ``\href{https://sketchfab.com/3d-models/oriental-building-f44115f33e954c75828d09cce03a729a}{Oriental Building}'' by N01516.
\item ``\href{https://sketchfab.com/3d-models/chinese-style-tea-table-3f0bae38d9a74ff6bf9ac5fcfb9fd176}{Chinese style tea table}'' by DailyArt.
\item ``\href{https://sketchfab.com/3d-models/porcelain-china-vase-656f83a4d9224c29abc82ade2207d2ce}{Porcelain China Vase}'' by rz.
\item ``\href{https://sketchfab.com/3d-models/chinese-lamp-dc117aeb4902472a952989314aa75d60}{Chinese lamp}'' by Coffeek.
\item ``\href{https://polyhaven.com/a/carrot_cake}{Carrot Cake}'' by Greg Zaal.
\item ``\href{https://polyhaven.com/a/painted_wooden_chair_02}{Painted Wooden Chair 02}'' by Kirill Sannikov.
\item ``\href{https://polyhaven.com/a/dutch_ship_large_01}{Dutch Ship Large 01}'' by James Ray Cock.
\item ``\href{https://polyhaven.com/a/marble_bust_01}{Marble Bust 01}'' by Rico Cilliers.
\item ``\href{https://polyhaven.com/a/brass_vase_03}{Brass Vase 03}'' by Rico Cilliers.
\item ``\href{https://polyhaven.com/a/jug_01}{Jug 01}'' by Kuutti Siitonen.
\item ``\href{https://polyhaven.com/a/ArmChair_01}{Arm Chair 01}'' by Kirill Sannikov.
\item ``\href{https://polyhaven.com/a/wooden_candlestick}{Wooden Candlestick}'' by Josh Dean.
\item ``\href{https://polyhaven.com/a/carved_wooden_elephant}{Carved Wooden Elephant}'' by Greg Zaal.
\item ``\href{https://polyhaven.com/a/pot_enamel_01}{Pot Enamel 01}'' by Kuutti Siitonen.
\item ``\href{https://polyhaven.com/a/rock_face_02}{Rock Face 02}'' by Dario Barresi.
\item ``\href{https://polyhaven.com/a/boulder_01}{Boulder 01}'' by Rico Cilliers.

\revise{
\item ``\href{https://sketchfab.com/3d-models/31e0899e65f74cb9bfdc9df8e71288ee}{ancient television on a table}'' by ricksticky.
\item ``\href{https://sketchfab.com/3d-models/61d5a28260744d8e8c4ba24d5d33db77}{Lowpoly viking helmet}'' by Dmytro Rohovyi.
\item ``\href{https://sketchfab.com/3d-models/afd8ddf280f14461945a18c4ab236859}{ballon}'' by Nyilonelycompany.
\item ``\href{https://sketchfab.com/3d-models/de0f1ffa93c64725a99e0f7a8b526501}{Woman's shirt}'' by Jacen Chio.
\item ``\href{https://sketchfab.com/3d-models/a7ca84f5b2224c9087aac4b4fb88ae3e}{Ratus}'' by dringoth.
\item ``\href{https://sketchfab.com/3d-models/0d32268905c645a4b41ba082af1c838a}{Octopus Clay Model [Re-upload]}'' by abot86.
\item ``\href{https://sketchfab.com/3d-models/b872e2c3f4f7457796edebcb7c0a290e}{Underwood 4-Bank Typewriter (Portable)}'' by Ed Swinbourne.
\item ``\href{https://sketchfab.com/3d-models/72960d59365844aebc8b9c85ae7b4cb7}{MOUNTAIN\_ BAKPOD©}'' by CIMORO.
\item ``\href{https://sketchfab.com/3d-models/7f996bcce4fd408f9c0db72cfd2e7e0e}{Turtle Project}'' by Scott Teel.
\item ``\href{https://sketchfab.com/3d-models/8bde00e0ca9540fba7d1d42ed76603e2}{Fishy}'' by steamsoldier.
\item ``\href{https://sketchfab.com/3d-models/571c5f9d65f142f1b3ee378196196283}{The Megaphone}'' by ezgi bakim.
\item ``\href{https://sketchfab.com/3d-models/ee5604b40e6740c29b5b67a384994d67}{sled pig}'' by maksimpetrik.
\item ``\href{https://sketchfab.com/3d-models/fa7f2be6b2b44aee936ea52b3c4b7330}{Low poly army boots}'' by tipicultbiomassa.
\item ``\href{https://sketchfab.com/3d-models/42c9bdc4d27a418daa19b2d5ff690095}{Stanford Bunny PBR}'' by hackmans.
\item ``\href{https://sketchfab.com/3d-models/d3f9aaecb7e94b12bc28256c85a40ce0}{Minotaur Statue}'' by plasmaernst.
\item ``\href{https://sketchfab.com/3d-models/84020334d1d047b8b77152108844e786}{Stagecoach}'' by Tuuttipingu.
\item ``\href{https://sketchfab.com/3d-models/b8d0e5ec21fe4650a723628600219142}{Cartoon Penguin WiP v1}'' by Drakahn Finlay.
\item ``\href{https://sketchfab.com/3d-models/c07ed2f916bc49ef8529043bbe8b7ca5}{Headphones Sony low}'' by danok98.
\item ``\href{https://sketchfab.com/3d-models/42cf842f793b462f87b35a24a88f608f}{Box}'' by KlGrimm.
\item ``\href{https://sketchfab.com/3d-models/90d32979dac545c6a465399e453c3484}{Aged Traffic Cone}'' by Eydeet.
\item ``\href{https://sketchfab.com/3d-models/0bef0065fbea496ca1caf3539c42e166}{Rep. 17}'' by SGMADEO\_1.

\item ``\href{https://sketchfab.com/3d-models/9f47a26ae86b4bb39695d3734dd78ee5}{\#powertool}'' by Digital Dressmaker.

\item ``\href{https://sketchfab.com/3d-models/cd3ec9bd9c76470c8ab72a93a86e946e}{Vintage Gold Pocket Watch}'' by Daz.
\item ``\href{https://sketchfab.com/3d-models/e6539a42c12a45488e196b14935cacf1}{Post apocalyptic style retro telephone}'' by Sousinho.
\item ``\href{https://sketchfab.com/3d-models/9a57589787144da49d7272fb153f726b}{Fan}'' by Escoly.
\item ``\href{https://sketchfab.com/3d-models/b39b032769ee4bdfbf42c137d5783806}{vaza}'' by protva2011.
\item ``\href{https://sketchfab.com/3d-models/8c3af1362caf45feb2e8eb2b6731926a}{Piano}'' by DarksProducer.
\item ``\href{https://sketchfab.com/3d-models/b6c404b9a7814023a4cccd91403e8472}{Biker}'' by KulerRuler.
\item ``\href{https://sketchfab.com/3d-models/20b9ef2a6579426da3a2806fdbb1c981}{Seashell 4K Photogrammetry | Game Ready asset}'' by Photogrammetry Guy.
\item ``\href{https://sketchfab.com/3d-models/32df7385c866435497f5a01954422429}{Damaged Leather Recliner}'' by Gravity Jack.
\item ``\href{https://sketchfab.com/3d-models/902ad095fca94981b6156fa49a21e495}{Dirtbike}'' by Thunder.
\item ``\href{https://sketchfab.com/3d-models/06d003d5a150421189b712eb876638c6}{Hand Painted SeaHorse}'' by stormk90.
\item ``\href{https://sketchfab.com/3d-models/7cce16ad3deb472086c9997e01884115}{Garbage can - Stylized}'' by Uricaro97.
}

\end{itemize}

Our style reference images are sourced from Civitai or directly generated using SD XL.
Style Images sourced from Civitai are attributed as follows:
\begin{itemize}
\item \href{https://civitai.com/models/51966/bohoai-konyconi} {BohoAI - konyconi}
\item \href{https://civitai.com/models/11203/glass-sculptures} {Glass Sculptures}
\item \href{https://civitai.com/models/62700/ivorygoldai-konyconi}{IvoryGoldAI - konyconi}
\item \href{https://civitai.com/images/10573118}{Glass mouse}
\item \href{https://civitai.com/models/190334/woodenmade}{Woodenmade}
\item \href{https://civitai.com/models/113008/doctor-diffusions-abstractor}{Doctor Diffusion's Abstractor}
\item \href{https://civitai.com/models/291/style-of-marc-allante}{style-of-marc-allante}
\item \href{https://civitai.com/images/6867307}{Ice cream}
\item \href{https://civitai.com/models/302929/pixel-particles-style-sdxl}{Pixel Particles Style [SDXL]}
\item \href{https://civitai.com/images/10183381}{Ink woman}
\item \href{https://civitai.com/models/225003/opal-style-lora-15sdxl}{Opal Style [LoRA 1.5+SDXL]
}
\item \href{https://civitai.com/models/234495/moss-beasts}{Moss Beasts}
\item \href{https://civitai.com/images/326147}{Anime Lineart / Manga-like Style}
\revise{
\item \href{https://civitai.com/models/126569/gentlecat-style}{GENTLECAT style}
\item \href{https://civitai.com/models/159674}{XL Realistic gold carving art style}
\item \href{https://civitai.com/models/231469/glacial-ice-style-sd15}{Glacial Ice Style [SD1.5]}

\item \href{https://civitai.com/models/288945/style-ceshi?modelVersionId=324908}{style ceshi}
\item \href{https://civitai.com/models/607019/style-of-milton-glaser-sdxl-368?modelVersionId=678572}{style of Milton Glaser [SDXL] 368}
\item \href{https://civitai.com/models/460531/style-of-raymond-duchamp-villon-sdxl-133}{style of Raymond Duchamp-Villon [SDXL] 133}
\item \href{https://civitai.com/models/153/style-darkestdungeon}{Style-Darkestdungeon}
\item \href{https://civitai.com/models/209978/ice-style-xl?modelVersionId=236506}{Ice Style XL}
\item \href{https://civitai.com/models/73305/zyd232s-ink-style}{zyd232's Ink Style}
\item \href{https://civitai.com/models/130868/pastel-color}{Pastel color}
\item \href{https://civitai.com/models/194825/xl-realistic-silver-carving-art-style}{XL Realistic silver carving art style}
\item \href{https://civitai.com/models/63565/necronomicon-pages}{Necronomicon Pages}
\item \href{https://civitai.com/models/120206/sdxlchinese-style-illustration}{(SDXL)chinese style illustration}
\item \href{https://civitai.com/models/67122/gelato-style?modelVersionId=71755}{Gelato Style}
\item \href{https://civitai.com/models/111408/niji-geometricshapes}{niji - geometric\_shapes}
\item \href{https://civitai.com/models/140979/schematics}{Schematics}
\item \href{https://civitai.com/models/135366/inkpunk-xl}{InkPunk XL}
}

\end{itemize}

\end{document}